\documentclass[11pt]{article}
\usepackage{jair, rawfonts}

\usepackage[colorlinks,urlcolor=blue,linkcolor=blue,citecolor=blue]{hyperref}
\usepackage{stmaryrd,scalerel}
\usepackage{color,array}
\usepackage{graphicx}
\usepackage{scalerel}
\usepackage{tikz}
\usetikzlibrary{svg.path}
\usepackage{xcolor}
\usepackage{amsmath}
\DeclareMathOperator*{\argmax}{argmax}
\DeclareMathOperator*{\argmin}{argmin}
\usepackage{amsfonts}
\usepackage{mathtools}
\usepackage[numbers]{natbib}
\usepackage[ruled,vlined,noend,linesnumbered]{algorithm2e}
\DontPrintSemicolon
\usepackage{multirow}
\usepackage{multicol}
\usepackage{booktabs}
\usepackage{blkarray}
\usepackage{fancyhdr}
\usepackage{float}
\usepackage[verbose]{placeins} 
\usepackage{wrapfig}
\usepackage{subcaption}

\definecolor{orcidlogocol}{HTML}{A6CE39}
\tikzset{
    orcidlogo/.pic={
        \fill[orcidlogocol] svg{M256,128c0,70.7-57.3,128-128,128C57.3,256,0,198.7,0,128C0,57.3,57.3,0,128,0C198.7,0,256,57.3,256,128z};
        \fill[white] svg{M86.3,186.2H70.9V79.1h15.4v48.4V186.2z}
        svg{M108.9,79.1h41.6c39.6,0,57,28.3,57,53.6c0,27.5-21.5,53.6-56.8,53.6h-41.8V79.1z M124.3,172.4h24.5c34.9,0,42.9-26.5,42.9-39.7c0-21.5-13.7-39.7-43.7-39.7h-23.7V172.4z}
        svg{M88.7,56.8c0,5.5-4.5,10.1-10.1,10.1c-5.6,0-10.1-4.6-10.1-10.1c0-5.6,4.5-10.1,10.1-10.1C84.2,46.7,88.7,51.3,88.7,56.8z};
    }
}

\newcommand\orcidicon[1]{\href{https://orcid.org/#1}{\mbox{\scalerel*{
                \begin{tikzpicture}[yscale=-1,transform shape]
                \pic{orcidlogo};
                \end{tikzpicture}
            }{|}}}}

\jairheading{}{}{}{}{}
\ShortHeadings{PAGE}
{Li \& Jha}

\begin{document}

\title{PAGE: Domain-Incremental Adaptation with Past-Agnostic Generative Replay for Smart Healthcare} 
\author{\name Chia-Hao Li $^{\textsuperscript{\orcidicon{0000-0001-9557-6050}}}$ \email chli@princeton.edu \\
        \addr Dept. of Electrical \& Computer Engineering, Princeton University\\
        Princeton, NJ 08544 USA\\
        \name Niraj K. Jha $^{\textsuperscript{\orcidicon{0000-0002-1539-0369}}}$, \textit{Fellow, IEEE} \email jha@princeton.edu \\
        \addr Dept. of Electrical \& Computer Engineering, Princeton University\\
        Princeton, NJ 08544 USA}
\maketitle

\begin{abstract}

Modern advances in machine learning (ML) and wearable medical sensors (WMSs) have enabled out-of-clinic disease detection. 
However, a trained ML model often suffers from misclassification when encountering non-stationary data domains after 
deployment. Use of continual learning (CL) strategies is a common way to perform domain-incremental adaptation while 
mitigating catastrophic forgetting. Nevertheless, most existing CL methods require access to previously learned domains 
through preservation of raw training data or distilled information. This is often infeasible in real-world scenarios 
due to storage limitations or data privacy, especially in smart healthcare applications. Moreover, it makes most 
existing CL algorithms inapplicable to deployed models in the field, thus incurring re-engineering costs. To address 
these challenges, we propose PAGE, a domain-incremental adaptation strategy with past-agnostic generative replay for 
smart healthcare. PAGE enables generative replay without the aid of any preserved data or information from prior domains. 
When adapting to a new domain, it exploits real data from the new distribution and the current model to generate 
synthetic data that retain the learned knowledge of previous domains. By replaying the synthetic data with the new 
real data during training, PAGE achieves a good balance between domain adaptation and knowledge retention. In addition, 
we incorporate an extended inductive conformal prediction (EICP) method into PAGE to produce a confidence score and a 
credibility value for each detection result. This makes the predictions interpretable and provides statistical 
guarantees for disease detection in smart healthcare applications. We demonstrate PAGE's effectiveness in 
domain-incremental disease detection with three distinct disease datasets collected from commercially available WMSs. 
PAGE achieves highly competitive performance against state-of-the-art along with superior scalability, data privacy, 
and feasibility. Furthermore, PAGE is able to enable up to 75\% reduction in clinical workload with the help of EICP.  \footnote[2]{\scriptsize This work has been submitted to the ACM for possible publication. Copyright may be transferred without notice, after which this version may no longer be accessible.\hfill} \footnote[3]{\scriptsize \copyright~2024 ACM. Personal use of this material is permitted. Permission from ACM must be obtained for all other uses, in any current or future media, including reprinting/republishing this material for advertising or promotional purposes, creating new collective works, for resale or redistribution to servers or lists, or reuse of any copyrighted component of this work in other works.\hfill}
\end{abstract}

\section{Introduction}
\label{sec:intro}
Physical and mental illnesses do not only affect our personal well-being but also impact global well-being. Especially 
after the overwhelming Coronavirus Disease 2019 (COVID-19) global pandemic, the world is seeking an efficient and 
effective strategy for disease detection. Thanks to the emergence of wearable medical sensors (WMSs) and modern 
advances in machine learning (ML), disease detection can now be made simple and accessible to not only health providers 
but also the general public, even in an out-of-clinic scenario \cite{covidD, diabD, mhD, ble_diab, arrhythmia, parkinson, 
MedAI, SAS}. However, a trained ML model often misclassifies when it encounters input data outside of the underlying 
probability distribution it was trained on. For example, an ML model trained to detect diabetes with data collected from 
the elderly might be inaccurate in detecting diabetes in young adults. Therefore, if an ML model cannot perform 
domain-incremental adaptation to retain its disease-detection accuracy in new data domains, it would place the
above-mentioned gains at risk.

Naively fine-tuning an ML model with data collected from new distributions degrades its performance on prior domains 
due to overfitting. This phenomenon in ML is well known as \textbf{\emph{catastrophic forgetting}} (CF) 
\cite{cataforget, review}. To alleviate CF, prior works have developed various continual learning (CL) strategies to 
perform domain-incremental adaptation while retaining learned knowledge \cite{overview, survey_a, survey_b, survey_c, 
survey_d}. These strategies include imposing restrictions on the update process of model parameters through 
regularization \cite{ewc, si, lwf}, dynamically reconfiguring model architectures to accommodate new tasks 
\cite{packnet, den, spacenet}, and replaying exemplars from past tasks during training \cite{icarl, remind, dgr, doctor}. 
However, most of the existing works focus on image-based CL tasks instead of nonstationary data distributions in 
tabular datasets. Tabular data are the most common data format for smart healthcare applications based on WMSs, such as 
physiological measurements and user demographics. Therefore, owing to the significant difference in their data 
modalities, the application of CL strategies to tabular data is still underexplored \cite{tabulardata, CL4Recurrent}. 

On the other hand, most existing CL strategies require access to previously learned domains through preservation of 
either raw training data or distilled information \cite{ewc, si, icarl, doctor}. Nevertheless, the use of a patient's 
personal health record is strictly regulated through numerous laws and regulations. Therefore, storing real patient 
data for exemplar replay in domain-incremental adaptation is often prohibited due to data privacy reasons. In addition, 
preserving raw data or distilled knowledge from all learned domains places a constraint on a system's scalability 
owing to storage limitations. Moreover, requesting stored data or information from prior CL domains may make a 
CL strategy inapplicable to existing models in the field. It confines most existing CL methods to re-architecting a 
new model from scratch (from the first CL task). However, this is often infeasible in real-world scenarios 
due to re-engineering costs.

To address the aforementioned challenges, we propose \textbf{PAGE}, a domain-incremental adaptation strategy with 
\textbf{\underline{p}}ast-\textbf{\underline{a}}gnostic \textbf{\underline{g}}enerative r\textbf{\underline{e}}play. 
PAGE targets WMS tabular data in smart healthcare applications, such as physiological signals and demographic 
information. It adopts a generative replay CL method that features past-agnostic synthetic data generation (SDG). When 
adapting to new domains, PAGE does not rely on replaying raw preserved data or exploiting distilled information from 
learned domains to mitigate CF. Instead, it takes advantage of real data from new domains to generate synthetic data 
that retain learned knowledge for replay. In addition, we incorporate an extended inductive conformal prediction (EICP) 
method in PAGE. EICP enables PAGE to generate a confidence score and a credibility value for each disease-detection 
result with little computational overhead. This enables model interpretability by providing users with statistical guarantees 
on its predictions. We summarized the advantages of PAGE next.

\begin{enumerate}
\item Scalability and privacy: PAGE has very low memory storage consumption since it does not store data or information 
from learned domains. Hence, PAGE is highly scalable to multi-domain disease detection while, at the same time,
preserving patient privacy. 
\item Feasibility: PAGE features a past-agnostic replay paradigm. This makes any off-the-shelf model 
amenable to domain-incremental adaptation, without the need to re-engineer new models from scratch. 
\item Clinical workload reduction: PAGE provides interpretability to disease-detection results. It enables users or 
medical practitioners to investigate if further clinician intervention is needed. When the model predictions are 
associated with high model confidence scores and credibility values, clinical workload can be significantly reduced. 
\end{enumerate}

The rest of the article is organized as follows. Section \ref{sec:back_rel} introduces related works on ML-driven 
disease detection based on WMSs and CL strategies, and provides background on probability density estimation and 
conformal prediction (CP). Section \ref{sec:def_metrics} presents problem definitions and evaluation metrics
employed in our work. Section \ref{sec:PAGE} gives details of our proposed strategy. Section \ref{sec:setup} provides 
information on the disease datasets used in our experiments and implementation details. Section \ref{sec:results} 
presents the experimental results. Section \ref{sec:dis_lim} discusses the limitations of our strategy and possible 
future work. Finally, Section \ref{sec:conclusion} concludes the article.

\section{Related Works and Background}
\label{sec:back_rel}

In this section, we discuss related works and background material that are helpful to understanding the rest of the
article.

\subsection{Machine Learning for Disease Detection with Wearable Medical Sensors}
The evolution of WMSs and ML has enabled efficient, effective, and accessible disease-detection systems. For example, 
Alfian \textit{et al.} \cite{ble_diab} propose a personalized healthcare monitoring system based on Bluetooth Low 
Energy (BLE)-based sensor devices and ML classifiers for diabetic patients. They gather distinct vital sign data from 
users with commercial BLE-based sensors. Then, the system analyzes the data with a multilayer perceptron (MLP) for 
early diabetes detection and a long short-term memory for future blood glucose level prediction. Farooq 
\textit{et al.} \cite{arrhythmia} introduce a cost-effective electrocardiography (ECG) sensor to obtain ECG
waveforms from patients and use K-means clustering to classify various arrhythmia conditions. Shcherbak 
\textit{et al.} \cite{parkinson} exploit commercial WMSs to collect motion data using designated exercises from test 
subjects and use ML classifiers to detect early-stage Parkinson's disease. Himi \textit{et al.} \cite{MedAI} construct 
a framework, called MedAi, that adopts a prototype smartwatch with eleven sensors to collect physiological signals from 
users and a random forest (RF) algorithm to predict numerous common diseases. Wang \textit{et al.} \cite{SAS} develop 
an RF-based diagnostic system to detect the sleep apnea syndrome by collecting human pulse wave signals and blood oxygen 
saturation levels with a photoplethysmography optical sensor. Hassantabar \textit{et al.} propose CovidDeep 
\cite{covidD} and MHDeep \cite{mhD} to detect the SARS-CoV-2 virus/COVID-19 disease and mental health disorders based 
on physiological signals collected with commercial WMSs and smartphones. They employ a grow-and-prune algorithm to 
synthesize optimal deep neural network (DNN) architectures for disease detection. Similarly, Yin \textit{et al.} build 
DiabDeep \cite{diabD} with grow-and-prune synthesis to deliver DNN models with sparsely-connected layers or 
sparsely-recurrent layers for diabetes detection based on WMS data and patient demographics. 

\subsection{Continual Learning Algorithms}
ML models suffer from performance degradation when presented with data from outside of the underlying domains they 
were trained on. Simply fine-tuning these models with new data results in CF. Therefore, researchers have been working 
on developing various algorithms to create more robust and general models. Domain adaptation \cite{DAReview, ADA, 
DA4Visual} is a sub-field in ML research that aims to address the aforementioned problem. However, most works in this 
sub-field focus on making ML models generalize from a source domain to a target domain. They are inapplicable to 
incremental multi-domain adaptation scenarios. Hence, we turn to CL strategies that enable ML models to perform 
domain-incremental learning (adaptation) in real-world scenarios with nonstationary data domains. Numerous CL algorithms 
have been proposed in the literature. Recent CL algorithms can mainly be categorized into three groups: 
regularization-based, architecture-based, and replay-based \cite{review, overview, survey_a, survey_b, survey_c, 
survey_d}. 

\subsubsection{Regularization-based Methods} 
These methods use distilled information for data-driven regularization to impose restrictions on the model 
parameter update process to prevent performance degradation on past tasks. A canonical example is \emph{elastic weight 
consolidation} (EWC) \cite{ewc}. EWC quadratically penalizes parameters that are important to previously-learned 
tasks, when deviating from the learned values. The importance of model parameters is weighted by the diagonal of a 
preserved Fisher information matrix distilled from prior tasks.

Another example in this category is \emph{synaptic intelligence} (SI) \cite{si}. SI introduces intelligence in each 
synapse to allow it to estimate its own importance in solving a given task based on its update trajectory. Then, 
SI stores this information as the weighting parameter for its quadratic regularization function to decelerate the 
parameter update process in order to mitigate CF on previous tasks.

Beyond restricting the update process of important parameters, \emph{learning without forgetting} (LwF) \cite{lwf} 
draws inspiration from knowledge distillation to perform CL with a multi-headed DNN architecture. Before learning a 
new task, LwF records output responses to new data features from each output head of the current network to generate 
various sets of pseudo labels. Pairs of pseudo labels and new data features distill learned knowledge of prior tasks 
from each output head. When LwF learns a new task, it trains on the new data features and their ground-truth labels 
on a new output head. In the meantime, it uses different sets of pseudo labels and copies of new data features to 
regularize the update process of their corresponding output heads to mitigate CF on past tasks.

Our work and LwF use a similar strategy to take advantage of new data and produce pseudo labels to perform CL. The 
main differences are that LwF aims to solve class-incremental learning problems with a multi-headed DNN and uses 
identical copies of new data features with their given pseudo labels for regularization. Directly applying LwF to 
domain-incremental adaptation problems with single-headed models can confuse the models since they are trained on the 
same set of data yet with different labels. In contrast, PAGE uses new real data to generate synthetic data and 
pseudo labels to perform generative replay to avoid this confusion in the training process. 

In summary, regularization-based methods address CF without storing raw exemplars but often require the preservation 
of distilled knowledge of prior tasks. Empirically, they do not achieve satisfactory performance under challenging CL
settings \cite{CLwVis} or complex datasets \cite{largeIL}. Thus, we do not adopt regularization-based CL algorithms 
in our strategy.

\subsubsection{Architecture-based Methods} 
These methods modify model architectures in various ways to accommodate new tasks while retaining learned knowledge. 
For example, PackNet \cite{packnet} interactively prunes the unimportant parameters after learning each task and 
retrains the spare connections for future tasks. In addition, it stores a parameter selection mask for each learned 
task to designate the connections reserved for the task at test time.

\emph{Dynamically expandable network} (DEN) \cite{den} learns new tasks by expanding the network capacity dynamically 
to synthesize a compact, overlapping, and knowledge-sharing structure. It uses a multi-stage process for CL. It first 
selects the relevant parameters from previous tasks to optimize. Then, it evaluates the loss on the new task after 
training. Once the loss exceeds a designated threshold, it expands network capacity and trains the expanded network 
with group sparsity regularization to avoid excessive growth.

On the other hand, SpaceNet \cite{spacenet} efficiently exploits a fixed-capacity sparse neural network. When learning 
a new task, it adopts an adaptive training method to train the network from scratch. This method reserves neurons that 
are important to learned tasks and produces sparse representations in the hidden layers for each learned task to reduce 
inter-task interference. Next, it effectively frees up unimportant neurons in the model to accommodate the new task. 

Architecture-based methods perform CL by expanding or sparing model capacity for new tasks. However, they might require 
a large number of additional parameters, be more complex to train, and not be scalable to many tasks. Hence, they 
generally are not favorable to wearable edge devices. Accordingly, we do not implement architecture-based CL methods in 
our framework either.

\subsubsection{Replay-based Methods} 
In general, replay-based methods store sampled exemplars from previous tasks in a small buffer or use generative models 
to generate synthetic samples that represent learned tasks. These samples are replayed when learning a new task to 
mitigate CF. For example, \emph{incremental classifier and representation learning} (iCaRL) \cite{icarl} selects 
groups of exemplars that best approximate their class centers in the latent feature space and stores them in a memory 
buffer. Then, it replays the stored samples with new data to perform nearest-mean-of-exemplars classification when 
learning a new class and at test time. In addition, it updates the memory buffer based on the distance between data 
instances in the latent feature space. 

Other than storing raw exemplars, \emph{replay using memory indexing} (REMIND) \cite{remind} imitates the hippocampal 
indexing theory by storing compressed data samples. It performs tensor quantization to efficiently store compressed 
representations of data instances from prior tasks for future replays. Therefore, it increases buffer memory efficiency 
and improves scalability of the framework. 

On the other hand, \emph{deep generative replay} \cite{dgr} uses a generative adversarial network \cite{gan} as the 
generator to generate synthetic images that imitate images from previous tasks. It uses a task-solving model as a 
solver to generate labels for the synthetic images to represent the knowledge of the learned tasks. When learning a 
new task, it interleaves the synthetic images and their labels with new data to update the generator and solver 
networks together. Generating synthetic data as a substitution for storing real data can preserve data privacy and 
allow the model to access as much training data as required for exemplar replay. 

Finally, DOCTOR \cite{doctor} is a state-of-the-art framework that targets various CL scenarios in the tabular data 
domain for smart healthcare applications. It exploits a replay-style CL algorithm that features an efficient exemplar 
preservation method and an SDG generative replay module. It allows users to choose between storing real samples or 
generating synthetic data for future replays. In addition, DOCTOR adopts a multi-headed DNN that enables simultaneous 
multi-disease detection. 

Generally, replay-based methods incur memory storage overhead to store exemplars and, hence, are not highly scalable. 
On the other hand, generative replay methods often require prior information to generate synthetic data. Therefore, it 
is challenging to apply these methods to existing models without re-engineering a new model from scratch. However, 
empirical results have demonstrated that they achieve the best performance trade-offs and are much stronger at 
alleviating CF than the other two methods, even when solving complex CL problems \cite{survey_c, tabulardata, 
perfectmemory}. Therefore, we develop a generative replay CL algorithm that requires no preserved tabular data or 
information from prior tasks.

\subsection{Probability Density Estimation}
\label{sec:para}
A function that describes the probability distribution of a continuous random variable is called a probability density 
function (PDF). It provides information about the likelihood of the random variable taking on a value within a specific 
interval. Probability density estimation methods are used to estimate the PDF of a continuous random variable based on 
observed data. These methods can be classified into two groups: parametric and non-parametric. In an ablation study 
presented in Section \ref{sec:ablation1}, we show that a parametric density estimation method performs better in our 
domain-incremental adaptation experiments with WMS data. Therefore, we implement the parametric density estimation method 
in PAGE. 

Parametric density estimation assumes that a PDF belongs to a parametric family. Therefore, density estimation is 
transformed into estimating various parameters of the family. One common example of parametric density estimation is 
the Gaussian mixture model (GMM). Suppose there are $N$ independent and identically distributed (\emph{i.i.d.}) samples 
$x_1, x_2, ..., x_N$ drawn from an unknown univariate distribution $f$ at any given point $x$. The estimated PDF of 
$f$ is represented as a mixture of $C$ Gaussian models in the form: 

\begin{equation*}
    p_X(x|\Theta) = \sum_{c=1}^Cp_{X|Z}(x|z_c,\Theta)\,p_Z(z_c|\Theta),
\end{equation*}

\noindent where $X$ is the observed variables, $\Theta$ represents the parameters of the GMM, $c$ is the Gaussian
model component, $z_c$ denotes the hidden state variable of $c$, and $Z$ symbolizes the hidden state variables that 
indicate the GMM assignment. The prior probability of each Gaussian model component $c$ can be written as:

\begin{equation*}
    p_Z(z_c|\Theta) = \theta_c.
\end{equation*}

\noindent Each Gaussian model component $c$ is a $d$-dimensional multivariate Gaussian distribution with mean vector 
$\mu_c$ and covariance matrix $\Sigma_c$ in the form:

\begin{equation*}
\begin{split}
    p_{X|Z}(x|z_c,\Theta) = &\left(\frac{1}{2\pi})\right)^{d/2}|\Sigma_c|^{-1/2}\exp\left(-\frac{1}{2}(x-\mu_c)^T\,\Sigma_c^{-1}\,(x-\mu_c)\right),
\end{split}
\end{equation*}

\noindent where $|\Sigma_c|$ is the determinant of the covariance matrix. Therefore, the PDF modeled by the GMM can 
be simplified as follows:

\begin{equation}
\label{eq:1}
    p_{X|C}(x) = \sum_{c=1}^C\pi_c\mathcal{N}(x|\mu_c,\Sigma_c),
\end{equation}

\noindent where $\pi_c$ represents the weight of the Gaussian model component $c$ and $\mathcal{N}$ denotes the normal distribution.

\subsection{Conformal Prediction}
\label{sec:CP}
CP was first proposed in \cite{CP1,CP2}. It complements each prediction result with a measure of confidence 
and credibility. The confidence score gives a measurement of the model uncertainty on the prediction. The credibility 
value indicates how suitable the training data are for producing the prediction. The paradigm can adopt any ML algorithm 
as the prediction rule, such as a support-vector machine, ridge regression, or a neural network \cite{CPTutorial}. By 
measuring how unusual a prediction result is relative to others, CP provides a statistical guarantee to a prediction 
through the confidence score and credibility value.

The original CP uses transductive inference to generate prediction with the underlying ML algorithm and all seen samples. 
However, it is highly computationally inefficient. Therefore, an alternative CP method based on inductive inference was 
proposed to address this issue \cite{ICP4NN, ICP4R, ICP4PR}. Here, we give an overview of the main concepts behind the 
transductive conformal prediction (TCP) and inductive conformal prediction (ICP) methods. For a more comprehensive 
description and statistical derivations, see \cite{CPTutorial,CP3}. 

\subsubsection{Transductive Conformal Prediction}
\label{sec:TCP}
For a classification task in a sample set $Z$, we are given a training set of examples $Z_{tr} = \{z_1, z_2, ..., z_n\} \subset Z$, where each example $z_i = (x_i, y_i)~|~i = 1, 2, ..., n \in Z_{tr}$ is a pair of a feature vector $x_i$ and its target label $y_i$. Our task is to predict the classification label $y_{n+1}$ for a new and unclassified vector $x_{n+1}$, where $z_{n+1} = (x_{n+1}, y_{n+1}) \in Z$. We know \textit{a priori} the set of all possible class labels 
$y^1, y^2, ..., y^c$, assume all samples in $Z$ are \emph{i.i.d.}. Under the \emph{i.i.d.} assumption, TCP provisionally 
iteratively assigns a class label $y^j~|~j = 1, 2, ..., c$ as the true label of the new vector $x_{n+1}$ and updates the 
underlying ML algorithm accordingly. Next, it measures the likelihood of class label $y^j$ being the true label of the 
new vector $x_{n+1}$ through a p-value function $p: Z^* \rightarrow [0,1]$. 

A p-value function gives the p-value of a class label assignment $y^j$ to the new sample $z_{n+1}^{(j)} = (x_{n+1}, y_{n+1} := y^j)$, denoted by $p(z_{n+1}^{(j)})$. A p-value function needs to satisfy the following properties \cite{ICP4NN, randomsequence}:

\begin{itemize}
    \item $\forall m \in \mathbb{N}, \forall \delta \in [0,1]$, and for all probability distributions $P$ on $Z$,
    \begin{equation*}
        P^m \{z \in Z^m: p(z) \leq \delta\} \leq \delta.
    \end{equation*}
    \item $p$ is semi-computable from above.
\end{itemize}

\noindent TCP constructs a p-value function by measuring how distinct each sample is from all others, namely measuring the \textit{non-conformity} of each sample. 

The non-conformity score of each sample is calculated by a family of non-conformity measures: 
$A_m: Z^{m-1} \times Z \rightarrow \mathbb{R}, \forall m \in \mathbb{N}$. Therefore, the non-conformity score of a sample $z_i$ can be formulated as:

\begin{equation*}
    \alpha_i = A_m(\lbag z_1, ..., z_{i-1}, z_{i+1}, ..., z_m \rbag, z_i)~|~i = 1, ..., m,
\end{equation*}

\noindent which indicates how different $z_i$ is from the other samples in the bag $\lbag z_1, ..., z_m \rbag \backslash \lbag z_i \rbag$. The non-conformity scores of all samples can then be used to compute the p-value of the class label assignment $y^j$ as:

\begin{equation*}
    p(z_{n+1}^{(j)}) := \frac{\# \{\alpha_{n+1}^{(j)} \leq \alpha_{i}~|~i = 1, ..., n+1\}}{n+1}.
\end{equation*}

After retrieving all p-values for all possible class labels $p(z_{n+1}^{(j)})~|~j = 1, ..., c$, TCP predicts the true label as the label with the largest p-value. Alongside, TCP gives a confidence score as one minus the second largest p-value and a credibility value as the largest p-value \cite{CPTutorial, ICP4NN}.

\subsubsection{Inductive Conformal Prediction}
\label{sec:ICP}
As mentioned in Section \ref{sec:TCP}, TCP iteratively updates the prediction rule with each possible label assignment $y^j$ to calculate its corresponding p-value. In other words, for any new feature vector $x_{n+r}, \forall r \in \mathbb{N}$ with $c$ possible class labels, TCP needs to retrain the underlying algorithm $c$ times. Next, TCP has to apply each updated prediction rule $n+r$ times to get the updated non-conformity scores for each seen sample to calculate the p-value of assignment 
$y^j$. This requires a significant amount of computations to perform one inference. Hence, it makes TCP highly unsuitable 
for lifelong learning and algorithms that require a long training period. 

To address the computational inefficiency, ICP was proposed in the literature \cite{ICP4NN, ICP4R, ICP4PR}. ICP adopts 
the same general ideas as TCP but uses a different strategy to compute the non-conformity measure. It splits the 
training set $Z_{tr}$ into two smaller sets: the proper training set $Z_{proper}$ with $p$ samples and the calibration 
set $Z_{calibrate}$ with $q := n-p$ samples. The proper training set is used to train the underlying ML algorithm and establish the prediction rule. The calibration set is used to calculate the p-values of each possible class label for any new sample $z_{n+r}$. More specifically, the non-conformity scores are calculated from the bag $\lbag z_{p+1}, ..., z_{p+q}, z_{n+r} \rbag$, which represents the degree of disagreement between the prediction rule built upon the bag $\lbag z_1, ..., z_p \rbag$ and the true labels $y_i~|~i = p+1, ..., p+q$ with the class assignment $y^j$ for $y_{n+r}$. This way, the prediction rule only needs to be trained once. Similarly, the non-conformity scores of the samples in the calibration set only need to be computed once as well. 

As a result, the p-value of each possible class label $y^j$ for $x_{n+r}$, with ICP, can be given as:

\begin{equation}
    \label{eq:2}
    p(z_{n+r}^{(j)}) := \frac{\# \{\alpha_{n+r}^{(j)} \leq \alpha_{i}~|~i = p+1, ..., p+q, n+r\}}{q+1}.
\end{equation}

\noindent Finally, ICP follows the same procedure as TCP to obtain the predicted label and generate its corresponding 
confidence score and credibility value, as described in Section \ref{sec:TCP}. 

\section{Problem Definitions and Evaluation Metrics}
\label{sec:def_metrics}

Next, we define the problem and present the metrics used for evaluation.

\subsection{Problem Definitions}
\label{sec:def}
CL is defined as training ML models on a series of tasks where non-stationary data domains, different classification 
classes, or distinct prediction tasks are presented sequentially \cite{overview, l2p}. Therefore, CL problems are 
usually categorized into three scenarios: domain-, class-, and task-incremental learning \cite{scenario}. In this work, 
we focus on the domain-incremental learning scenario for our disease-detection tasks. 

Domain-incremental adaptation can be defined as the scenario where data from different probability distribution domains 
for the same prediction task become available sequentially. For example, data collected from different cities, countries, 
or age groups may become available incrementally for the same disease-detection task. Given a prediction task, we define 
a potentially infinite sequence of various domains as $D = \{D_1, D_2, ..., D_n, ...\}$, where the $n$-th domain is depicted by $D_n = \{(X_n^i, Y_n^i)~|~i = 1, 2, ..., C\}$. $X_n^i \in \mathcal{X}$ and $Y_n^i \in \mathcal{Y}$ refer to the set of data features and their corresponding target labels for the $i$-th class in domain $D_n$. $C$ refers to the total number of classification classes for the prediction task. The objective of domain-incremental adaptation is to train an ML model $f_\theta: \mathcal{X} \rightarrow \mathcal{Y}$, parameterized by $\theta$, such that it can incrementally learn new data domains in $D$ and predict their labels $Y = f_\theta(X \in \mathcal{X}) \in \mathcal{Y}$ without degrading performance on prior 
domains. 

\subsection{Evaluation Metrics}
\label{sec:metrics}
Several metrics are commonly used to evaluate CL algorithms in the literature, including average accuracy, average
forgetting, backward transfer (BWT), and forward transfer \cite{overview, survey_d}. We report the average accuracy
and average F1-score across all learned domains in our experimental results to validate PAGE's effectiveness in
domain-incremental adaptation for various disease-detection tasks. When reporting the average F1-score, we define
true positives (negatives) as the unhealthy (healthy) instances correctly classified as disease-positive (healthy)
and false positives (negatives) as the healthy (unhealthy) instances misclassified as disease-positive (healthy).
In addition, we report BWT to evaluate how well PAGE mitigates CF. BWT measures how much a CL strategy impacts a 
model's performance on previous tasks when learning a new one. For a model that has learned a total of $q$ tasks, 
BWT can be defined as:

\begin{equation*}
    \text{BWT} = \frac{1}{q-1}\sum_{n=1}^{q-1}(a_n^q - a_n^n),
\end{equation*}

\noindent where $a_n^q$ represents the model's test accuracy for the $n$-th task after continually learning $q$ tasks, and $a_n^n$ denotes the model's test accuracy for the $n$-th task after continually learning $n$ tasks.

\section{The Past-Agnostic Generative-Replay Strategy}
\label{sec:PAGE}

We detail our methodology next.

\begin{figure}[t]
    \centering
    \includegraphics[width=1\linewidth]{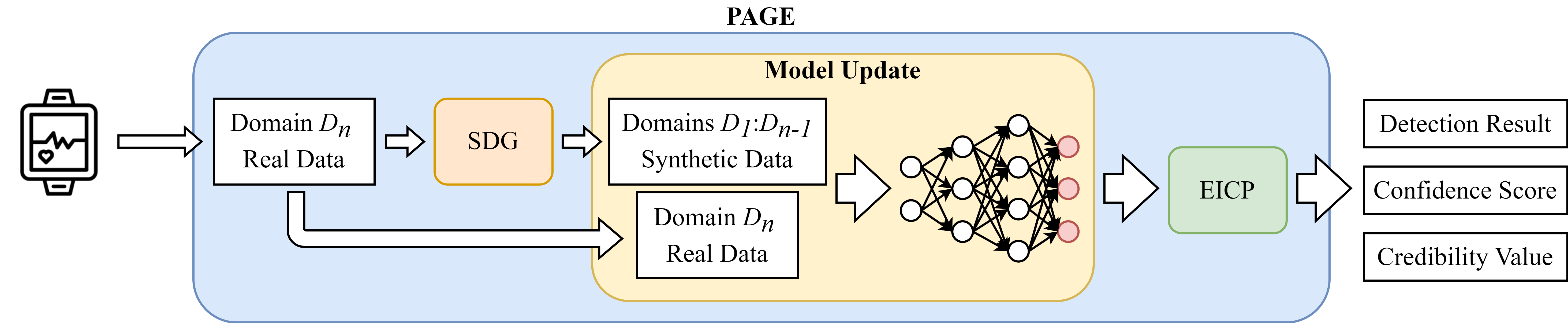}
    \caption{The schematic diagram of our PAGE strategy.}
    \label{fig:schematic}
\end{figure}

\subsection{Strategy Overview}
First, we provide an overview of the proposed PAGE strategy. Fig.~\ref{fig:schematic} shows its schematic diagram. 
It is composed of three steps for performing domain-incremental adaptation and generating system outputs. For example, 
we use PAGE to learn a new data domain $D_n$ with an ML model that has learned $n-1$ domains for a disease-detection task 
using WMS data. PAGE passes the new real WMS data from $D_n$ through an SDG module to generate synthetic data that 
retains the knowledge learned from domains $D_1$ to $D_{n-1}$. Next, the synthetic data are replayed together with the 
new real data to update the model in a generative replay fashion. Finally, PAGE uses the proposed EICP method to produce 
detection results and their corresponding confidence scores and credibility values. 

\begin{figure*}[t!]
    \centering
    \begin{subfigure}[b]{0.276\textwidth}
        \centering
        \includegraphics[width=\textwidth]{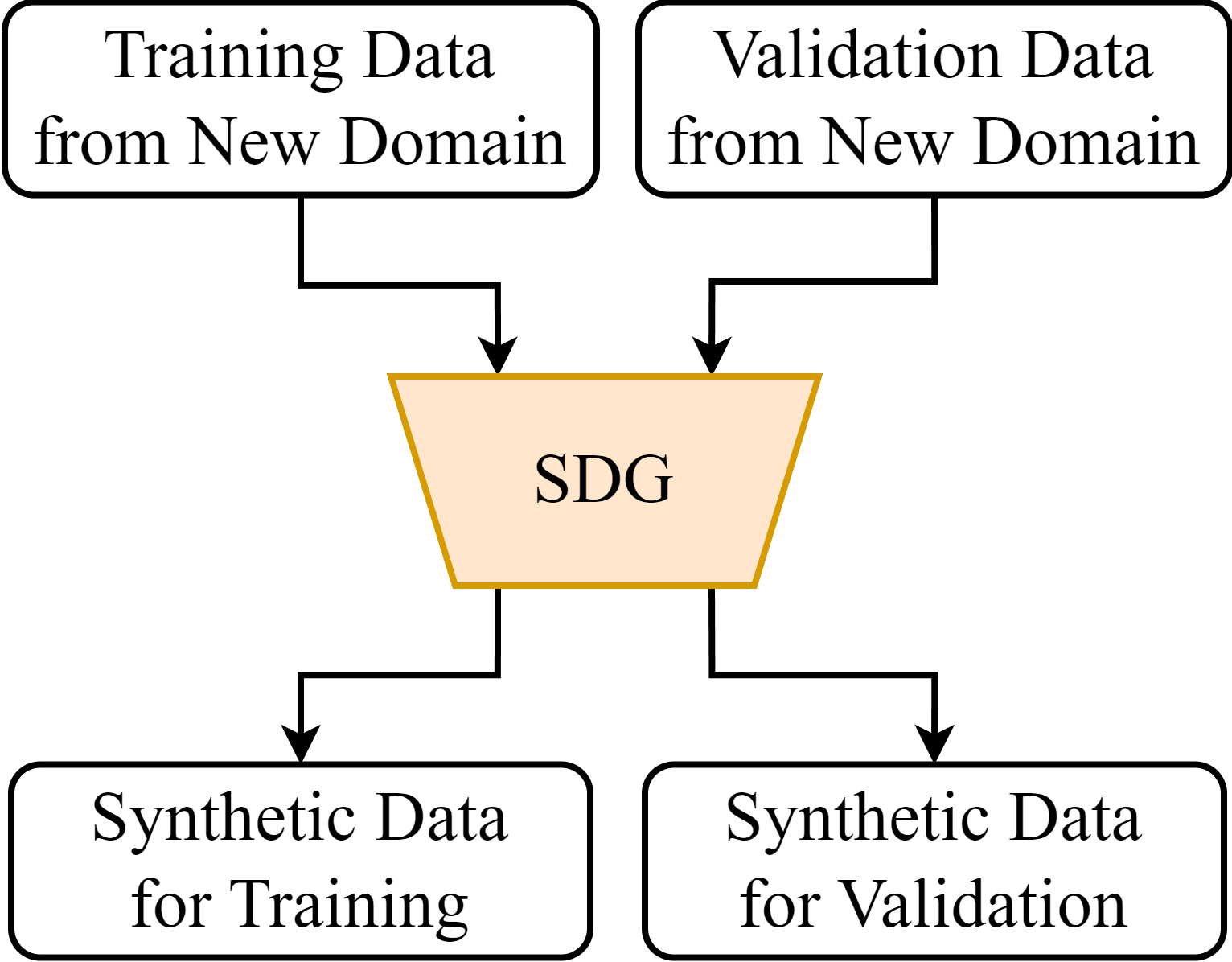}
        \caption{Step 1}
        \label{fig:flow1}
    \end{subfigure}
    \begin{subfigure}[b]{0.281\textwidth}
        \centering
        \includegraphics[width=\textwidth]{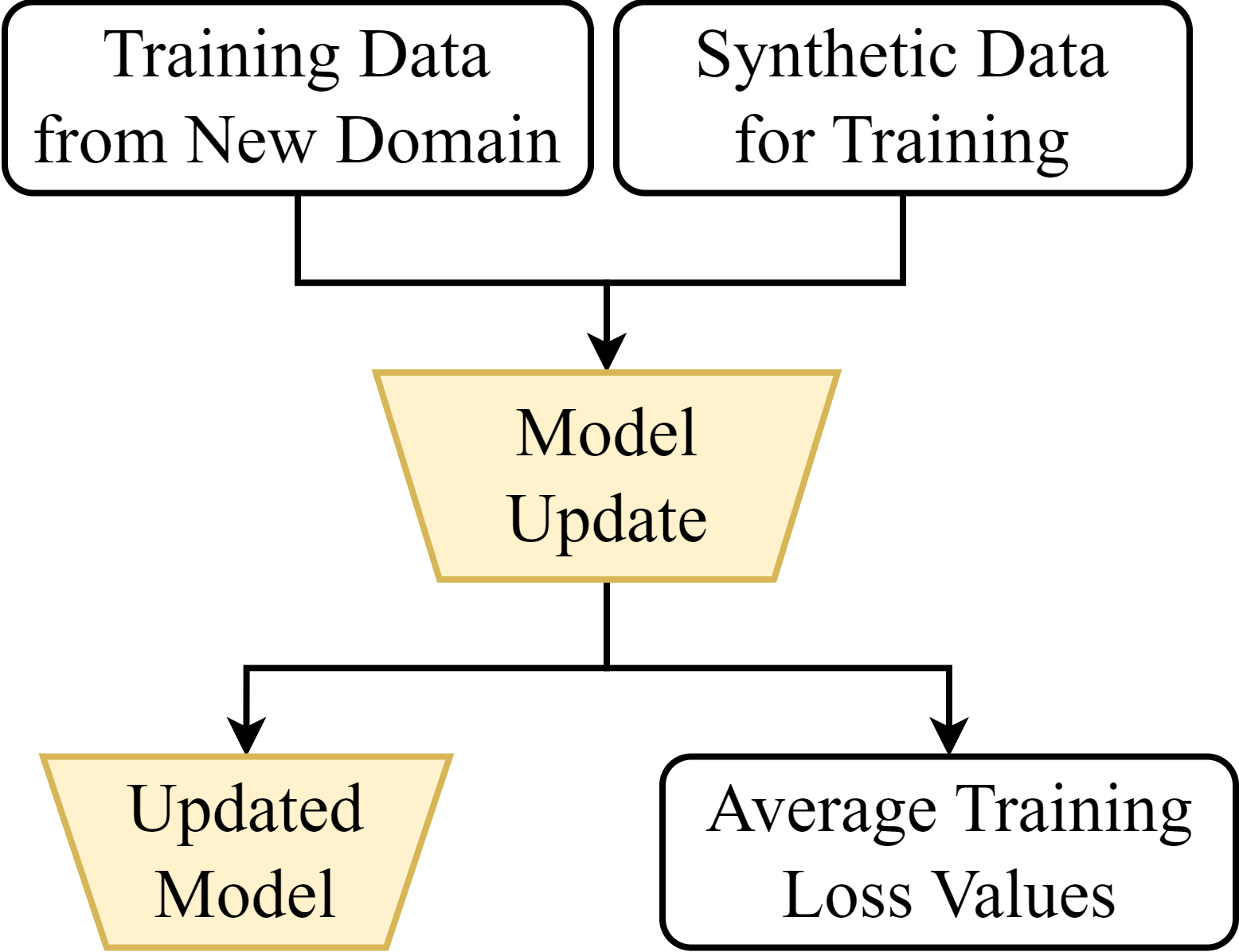}
        \caption{Step 2}
        \label{fig:flow2}
    \end{subfigure}
    \begin{subfigure}[b]{0.42\textwidth}
        \centering
        \includegraphics[width=\textwidth]{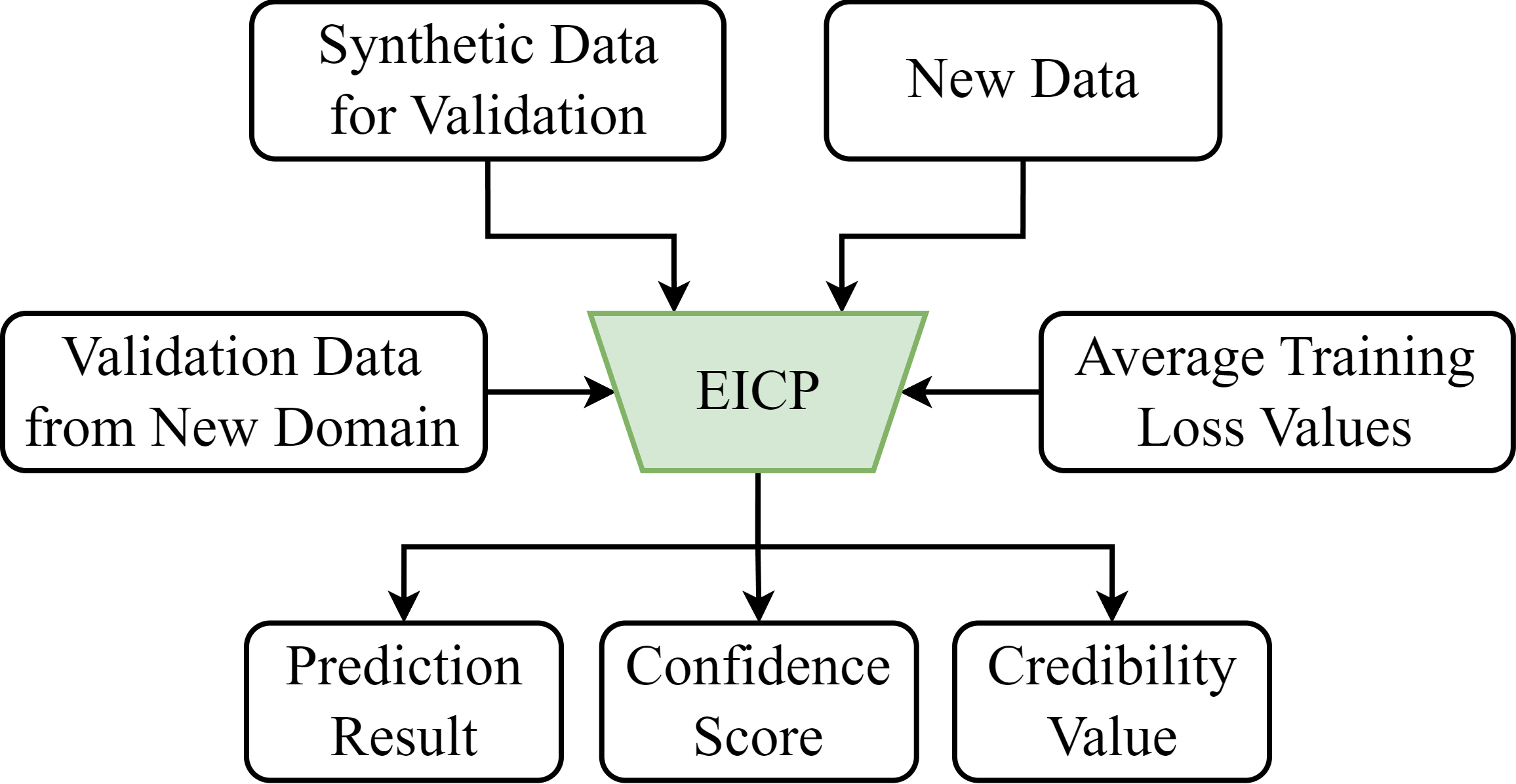}
        \caption{Step 3}
        \label{fig:flow3}
    \end{subfigure}
    \caption{The top-level flowchart of our PAGE strategy.}
    \label{fig:flowchart}
\end{figure*}

Fig.~\ref{fig:flowchart} shows the top-level flowchart of the three steps of PAGE. In Step 1, as shown in 
Fig.~\ref{fig:flow1}, the SDG module exploits only the real training and validation data from the new domain to produce 
synthetic data for training and validation, respectively. In Step 2, depicted in Fig.~\ref{fig:flow2}, the 
synthetic data for training are replayed with the real training data from the new domain to update the model. During 
this process, PAGE also records the average training loss values of \emph{all} training data for Step 3. In 
Step 3, illustrated in Fig.~\ref{fig:flow3}, EICP uses the real validation data from the new domain, the 
synthetic data generated for validation, the average training loss values, and the new data for detection as inputs. 
Finally, EICP generates detection results along with their corresponding confidence scores and credibility values to 
provide model interpretability and statistical guarantees to users. 

Next, we dive into the details of each step of PAGE.

\subsection{Synthetic Data Generation}
\label{sec:SDG}

The SDG module generates synthetic data that include pairs of synthetic data features and pseudo labels for replay, to 
retain knowledge of learned domains. We draw inspiration from a framework called TUTOR \cite{tutor} to build our SDG 
module. First, the module models the joint multivariate probability distribution of the real data features from the 
new domain through a probability density estimation function. Next, it generates synthetic data features by sampling 
from the learned distribution. Then, it passes the synthetic data features to the current model and records the 
responses from the output softmax layer. It assigns these raw output probability distributions as pseudo labels of 
the synthetic data features. The pairs of synthetic data features and their pseudo labels effectively represent the 
knowledge of the previous domains that the model has learned so far. Therefore, we can replay the synthetic data 
during the model update process to retain the learned knowledge.

We adopt a multi-dimensional GMM as the probability density estimation function in our SDG module. The probability 
distribution of the real data features is modeled as a mixture of $C$ Gaussian models, as mentioned in Section 
\ref{sec:para}. The PDF of the learned distribution can be expressed using Eq.~(\ref{eq:1}). Note that previous works 
\cite{covidD, mhD, tutor, repairs} have already shown that GMM can recreate the probability distribution of real data 
and generate high-quality synthetic data. The generated synthetic data can then replace real data in the training process 
while achieving similar test accuracy. However, one drawback of this method lies in its learning complexity. To address 
this problem, we use the expectation-maximization (EM) algorithm to determine the GMM parameters \cite{tutor}. The 
EM algorithm iteratively solves this optimization problem in the following two steps until convergence:

\begin{itemize}
\item Expectation-step: Given the current parameters $\theta_i$ of the GMM and observations $X$ from the real training 
data, estimate the probability that $X$ belongs to each Gaussian model component $c$.
\item Maximization-step: Find the new parameters $\theta_{i+1}$ that maximize the expectation derived from the 
expectation-step.
\end{itemize}

Algorithm \ref{alg:SDG} shows the pseudocode of the SDG module. It is important to note that the input real data
features for training GMM $X_{train\_gmm}$ and the input real data features for validating GMM $X_{valid\_gmm}$
refer to the real data features that are \emph{used to train and validate the underlying GMM}, respectively. They
\emph{do not} stand for the real training data features $X_{train}$ and real validation data features $X_{valid}$
from the new domain. The algorithm starts by initializing a set $\mathcal{C}$ of Gaussian model components, ranging 
from 1 to a user-specified maximum number $C_{max}$. Then, for each candidate number $C \in \mathcal{C}$, the algorithm 
fits a GMM $gmm_C$ to the given $X_{train\_gmm}$ with $C$ Gaussian model components. To determine the optimal total 
number of components $C^*$, we use the Bayesian information criterion (BIC) \cite{BIC} for model selection. BIC is a 
commonly used criterion in statistics to measure how well a model fits the underlying data. BIC favors models that have 
a good fit with fewer parameters and smaller complexity, which results in a smaller chance of overfitting 
\cite{modelselectreview}. The algorithm computes the BIC score $\texttt{bic()}$ of each GMM $gmm_C$ to evaluate the 
quality of the model. To further avoid the risk of overfitting, we use this criterion with the given $X_{valid\_gmm}$. A 
lower BIC score implies a better fit in model selection. Therefore, the number $C$ that minimizes this criterion is 
chosen as the optimal $C^*$. This means that $C^*$ Gaussian models are required to model the probability distribution 
of $X_{train\_gmm}$. Then, the algorithm finalizes the optimal GMM $gmm^*$ with $C^*$ components to fit 
$X_{train\_gmm}$. Next, it samples a user-defined number $count$ of synthetic data features $X_{syn}$ from $gmm^*$. 
Finally, we pass $X_{syn}$ to the current model $\mathbf{M}$ and assign the raw output probability distributions from 
its softmax output layer as their pseudo labels $Y_{syn}$. 

\SetKwComment{Comment}{\# }{}
\begin{algorithm}[t]
\small
\SetAlgoLined
    \SetKwInOut{Input}{Input}
    \SetKwInOut{Output}{Output}
    \Input{current model ($\mathbf{M}$), real data features for training GMM ($X_{train\_gmm}$), real data features
for validating GMM ($X_{valid\_gmm}$), maximum number of components ($C_{max}$), number of synthetic data instances ($count$)}
    \Output{synthetic data features ($X_{syn}$), pseudo labels ($Y_{syn}$)}
    \BlankLine
    initialize a set of component numbers $\mathcal{C} = \texttt{range}(1, C_{max}, \texttt{step}=1)$\;
    \ForEach {$C \in \mathcal{C}$}{
        $gmm_C = \texttt{GMM.fit}(C, X_{train\_gmm})$ \Comment*{using EM algorithm}
    }
    $C^* = \argmin\limits_{\mathcal{C}}(gmm_C \texttt{.bic}(X_{valid\_gmm}))$\;
    $gmm^* = \texttt{GMM.fit}(C^*, X_{train\_gmm})$\;
    $X_{syn} = gmm^*\texttt{.sample}(count)$\;
    $Y_{syn} = \mathbf{M}(X_{syn})$ \Comment*{probability distribution from the softmax output layer}
    \caption{Synthetic Data Generation}
    \label{alg:SDG}
\end{algorithm}

We repeat this procedure twice to generate synthetic data for training and validation. When we generate the synthetic 
data for training $(X_{syn\_train}, Y_{syn\_train})$, the inputs are $X_{train\_gmm} = X_{train}$ and 
$X_{valid\_gmm} = X_{valid}$. When we generate the synthetic data for validation $(X_{syn\_valid}, Y_{syn\_valid})$, 
the inputs are $X_{train\_gmm} = X_{valid}$ and $X_{valid\_gmm} = X_{train}$. 

\subsection{Model Update}
\label{sec:MU}

\SetKwComment{Comment}{\# }{}
\begin{algorithm}[t]
\small
\SetAlgoLined
    \SetKwInOut{Input}{Input}
    \SetKwInOut{Output}{Output}
    \SetKwInOut{Parameter}{Parameter}
    \Input{real training data features ($X_{train}$), real training data labels ($Y_{train}$), real validation data features ($X_{valid}$), real validation data labels ($Y_{valid}$), synthetic training data features ($X_{syn\_train}$), pseudo labels for training ($Y_{syn\_train}$), synthetic validation data features ($X_{syn\_valid}$), pseudo labels for training ($Y_{syn\_valid}$), number of epochs ($E$), mini-batch size ($B$)}
    \Output{updated model ($\mathbf{M}_{new}$), vector of average trailing loss values ($\mathcal{L}$)} 
    \Parameter{model parameters ($\theta$)}
    \BlankLine
    initialize the vector of average training loss values $\mathcal{L}$\;
    $x_{valid}$ = \texttt{concatenate}($X_{valid}$, $X_{syn\_valid}$)\;
    $y_{valid}$ = \texttt{concatenate}($Y_{valid}$, $Y_{syn\_valid}$)\;
    \For {e = 1 to E}{
        initialize a temporary vector $v$\;
        \While {not exhausted instances in $X_{train}$ and $X_{syn\_train}$}{
            $x_{real}$ = samples from $X_{train}$ of size $\frac{B}{2}$\;
            $y_{real}$ = labels of $x_{real}$ from $Y_{train}$\;
            $x_{syn}$ = samples from $X_{syn\_train}$ of size $\frac{B}{2}$\;
            $y_{syn}$ = labels of $x_{syn}$ from $Y_{syn\_train}$\;
            $x_{train}$ = \texttt{concatenate}($x_{real}$, $x_{syn}$)\;
            $y_{train}$ = \texttt{concatenate}($y_{real}$, $y_{syn}$)\;
            update $\theta$ with ($x_{train}$, $y_{train}$)\;
            \ForEach{instance in ($x_{train}$, $y_{train}$)}{
                append the instance's training loss in $v$\;
            }
        }
        $\mathcal{L}$ += $v$\;
        store $\theta$ and their performance on ($x_{valid}, y_{valid}$)\;
    }
    $\mathbf{M}_{new}$ $\leftarrow$ update the model with the $\theta$ that maximizes the performance on ($x_{valid}, y_{valid}$)\;
    $\mathcal{L} = \mathcal{L} / E$ \Comment*{obtain the average training loss values over $E$ epochs}
    \caption{Model Update} 
    \label{alg:MU}
\end{algorithm}

In this step, we update the model and record the average training loss values of all the training data, 
real and synthetic, for the next step. To update the model while mitigating CF, we replay the synthetic training data 
with the real training data from the new domain during training. Due to the difference in dataset sizes between the 
real and synthetic training data, the model might be biased toward the dataset with more data. To avoid the biased 
prediction problem, we sample the same number of data instances from each dataset to form each mini-batch to train the 
model.  In other words, if we set the mini-batch size to $B$, $\frac{B}{2}$ of the data are sampled from the real 
training data from the new domain while the other $\frac{B}{2}$ of the data are sampled from the synthetic training data. 

Algorithm \ref{alg:MU} demonstrates the pseudocode of the model update procedure. In the beginning, we initialize a 
vector $\mathcal{L}$ to store the average training loss of each data instance used in the training process. 
In addition, we concatenate the real and synthetic validation data features ($X_{valid}, X_{syn\_valid}$) and their 
labels ($Y_{valid}, Y_{syn\_valid}$) to prepare validation data ($x_{valid}$, $y_{valid}$) for model update. Then,
we train the model for $E$ epochs with training data in a mini-batch size $B$. Within each epoch, while data
instances are not exhausted in both real training data ($X_{train}$, $Y_{train}$) and synthetic training data
($X_{syn\_train}$, $Y_{syn\_train}$), we sample $\frac{B}{2}$ data instances from ($X_{train}$, $Y_{train}$) and 
another $\frac{B}{2}$ from ($X_{syn\_train}$, $Y_{syn\_train}$) to form a mini-batch data ($x_{train}$, $y_{train}$) 
to update the model parameters $\theta$. For each data instance in ($x_{train}$, $y_{train}$), we record the training 
loss value in a temporary vector $v$. At the end of each epoch, we accumulate the training loss values of all training 
data instances in vector $\mathcal{L}$. Also, we store the current $\theta$ and the resulting performance on 
($x_{valid}$, $y_{valid}$). After updating the model for $E$ epochs, we assign the $\theta$ that maximizes performance 
on ($x_{valid}$, $y_{valid}$) to the updated model $\mathbf{M}_{new}$. Finally, we obtain the average of the training 
loss values over $E$ epochs to retrieve the vector of average training loss values $\mathcal{L}$ for all training data. 

\begin{figure*}[t!]
    \centering
    \begin{subfigure}[b]{0.253\textwidth}
        \centering
        \includegraphics[width=\textwidth]{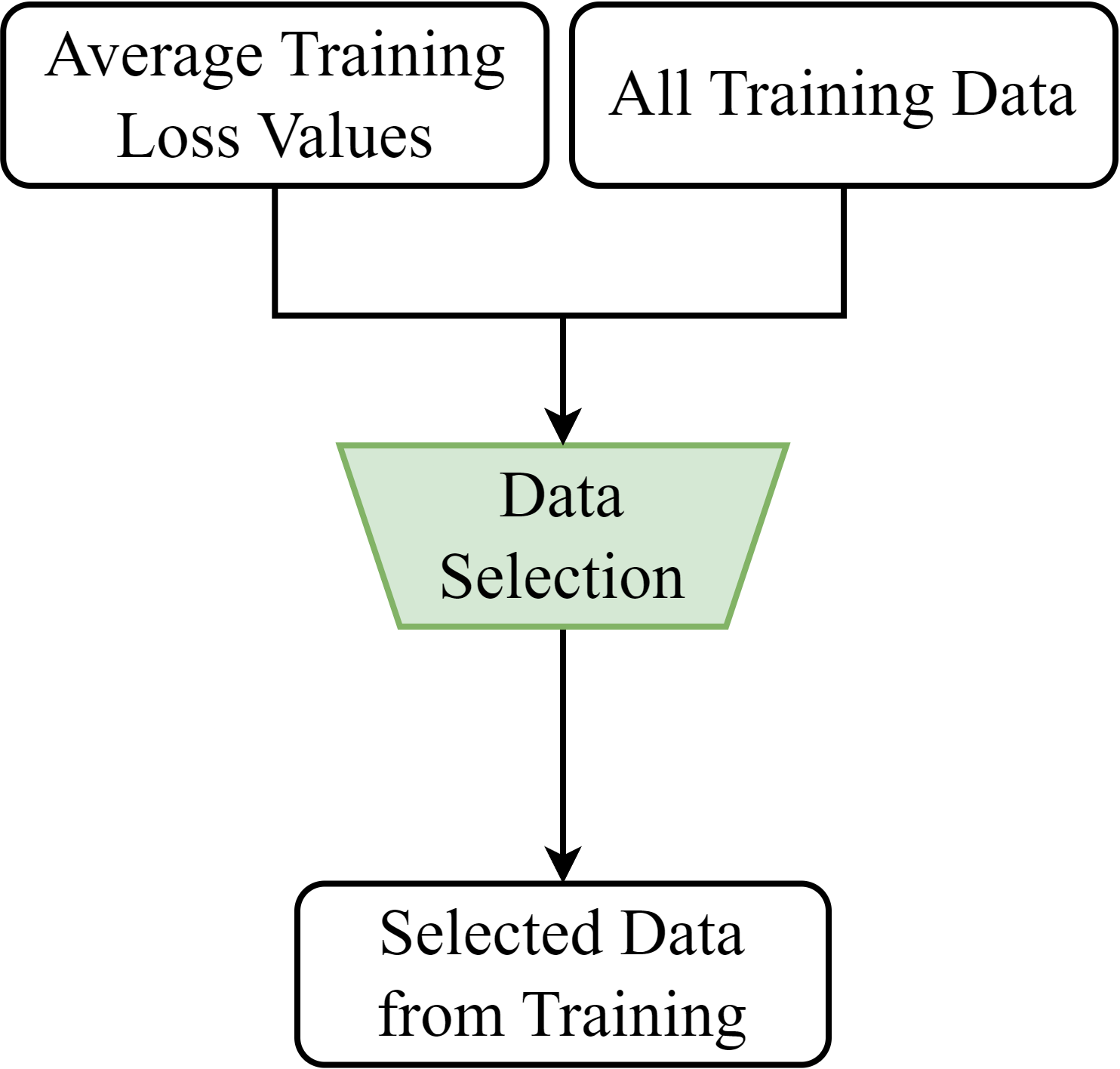}
        \caption{Process 1}
        \label{fig:eicp1}
    \end{subfigure}
    \begin{subfigure}[b]{0.38\textwidth}
        \centering
        \includegraphics[width=\textwidth]{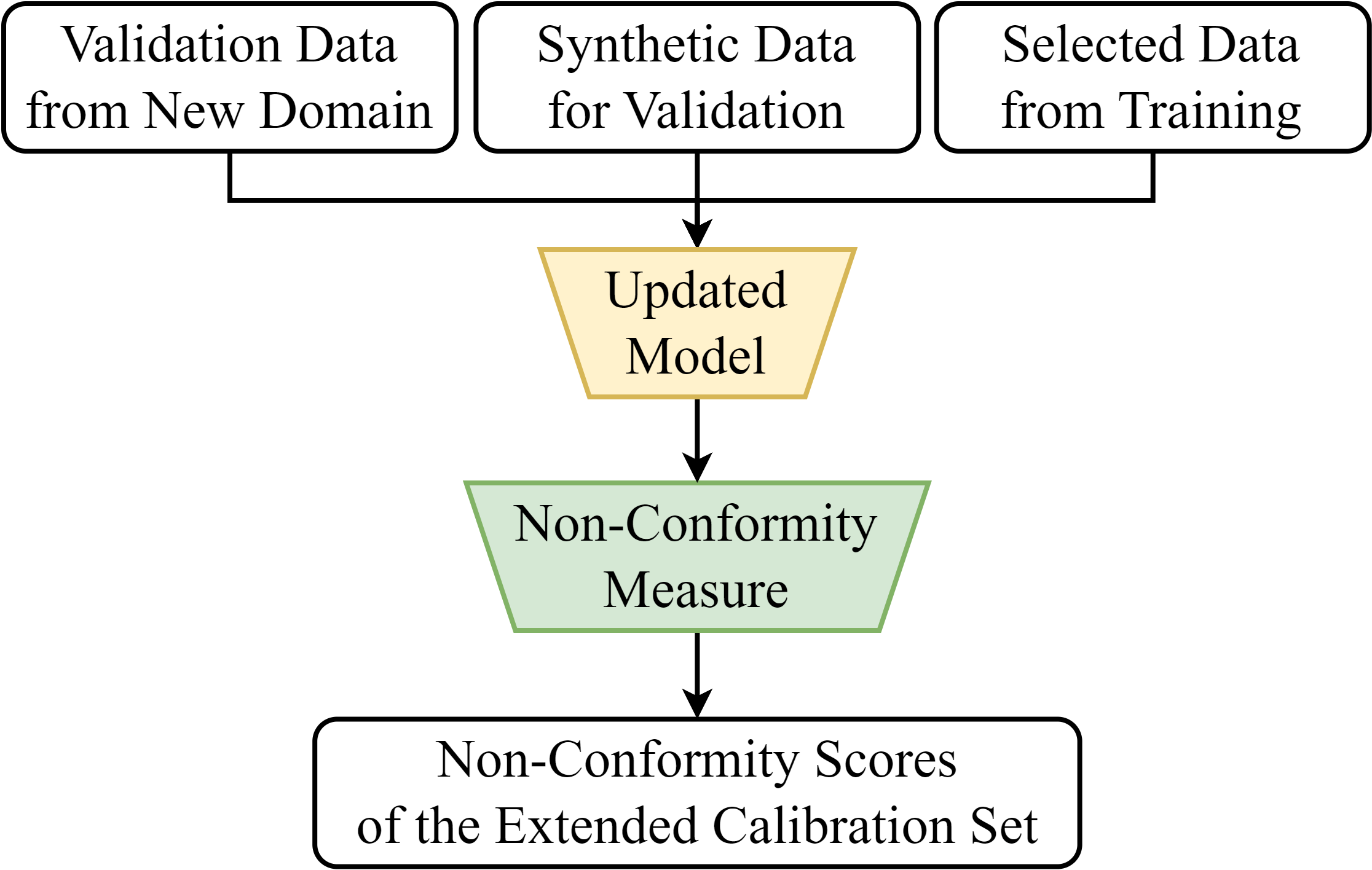}
        \caption{Process 2}
        \label{fig:eicp2}
    \end{subfigure}
    \begin{subfigure}[b]{0.34\textwidth}
        \centering
        \includegraphics[width=\textwidth]{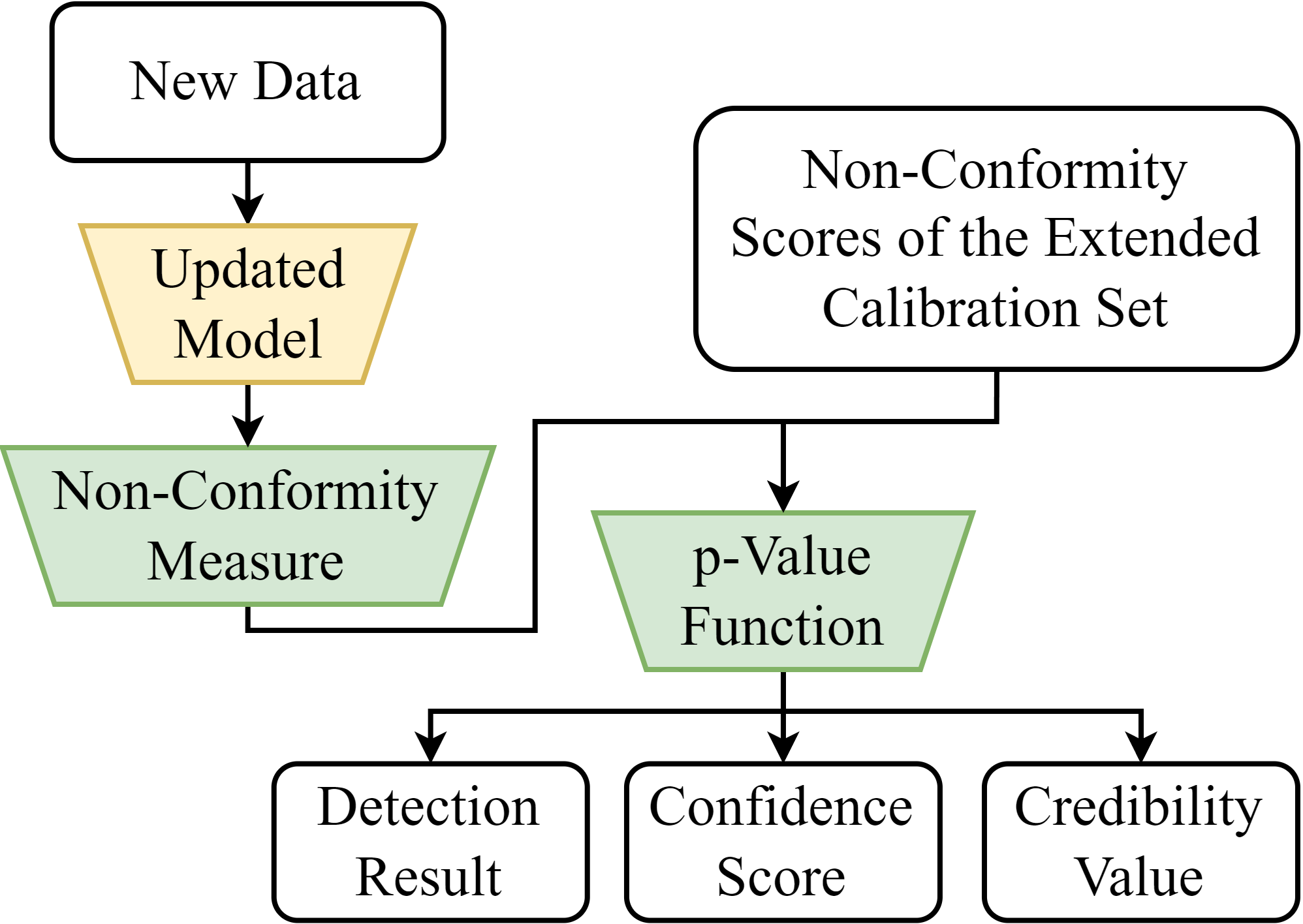}
        \caption{Process 3}
        \label{fig:eicp3}
    \end{subfigure}
    \caption{The flowchart of EICP.}
    \label{fig:eicp}
\end{figure*}

\subsection{Extended Inductive Conformal Prediction}
\label{sec:EICP}
Next, we propose EICP, an extended version of the original ICP, to generate a detection result, confidence score, and 
credibility value for a new data instance. The new data instance can belong to any domain that the model has learned 
so far. Similar to the original ICP introduced in Section \ref{sec:ICP}, we combine the real and synthetic training 
data into a proper training set to train the underlying ML model. Next, we use the real and synthetic validation data 
as a calibration set to calculate non-conformity scores. As opposed to ICP, we extend the calibration set by including 
a subset of the proper training set. By doing so, we strengthen the ability of the calibration set to perform CP 
based on the learned domains. 

Fig.~\ref{fig:eicp} shows the flowchart of EICP. It consists of three processes. In Process 1, EICP selects a subset 
of data from all training data, including the real and synthetic training data, to form the extended calibration set. 
As shown in Fig.~\ref{fig:eicp1}, EICP selects the data through a data selection (DS) module based on their average 
training loss values. In Process 2, EICP computes the non-conformity scores of the extended calibration set through 
a non-conformity measure. As shown in Fig.~\ref{fig:eicp2}, the softmax output responses of the extended calibration 
set from the updated model are used to compute their non-conformity scores. In Process 3, EICP first retrieves the 
softmax output response of the new data instance from the updated model and computes its non-conformity score using
the non-conformity measure. Finally, as illustrated in Fig.~\ref{fig:eicp3}, EICP produces the detection result for
the new data instance, confidence score, and credibility value using the non-conformity scores of the new data instance 
and the calibration set through a p-value function, as described in Section \ref{sec:ICP}.

\begin{figure}[t]
    \centering
    \includegraphics[width=0.6\linewidth]{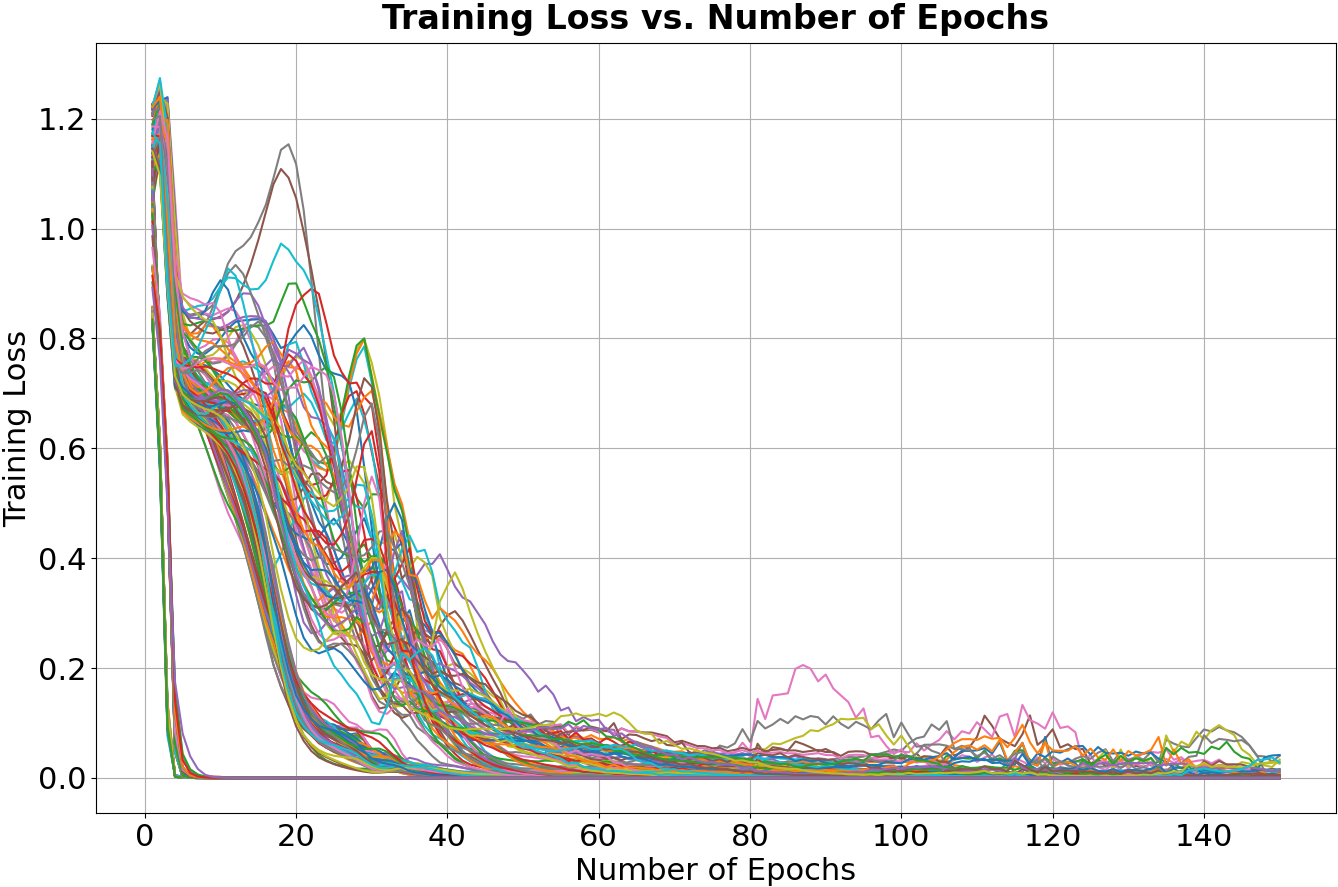}
    \caption{The training loss of each training data instance in the CovidDeep dataset \cite{covidD} across 150 epochs.}
    \label{fig:loss}
\end{figure}

\subsubsection{Data Selection}
\label{sec:DS}
We take inspiration from a framework called \emph{clustering training losses for label error detection} (CTRL) 
\cite{ctrl} to construct the DS module. We look at the average training loss of each training data instance to evaluate 
how suitable the instance is for selection in the extended calibration set. Fig.~\ref{fig:loss} presents a graph of 
the training loss curves of the real training data in the CovidDeep dataset \cite{covidD} across 150 epochs. The 
training loss values of some data instances decrease fast and stay near zero after around 30 epochs. As a result, 
these data instances have low average training loss values. They represent the easier-to-learn information in the 
given domain and, hence, are less non-conformal compared to other data. On the other hand, the training loss values 
of some data remain high until around 70 epochs. These data instances yield higher average training loss values and 
require more epochs for the model to learn. Therefore, they represent the difficult-to-learn knowledge in the underlying 
domain. These data are good candidates for helping EICP calibrate the non-conformity of a new data instance and its 
provisional class label assignment against the learned knowledge of the model. We conduct an ablation study for 
this design decision and give details in Section \ref{sec:ablation3}.

\SetKwComment{Comment}{\# }{}
\begin{algorithm}[t]
\small
\SetAlgoLined
    \SetKwInOut{Input}{Input}
    \SetKwInOut{Output}{Output}
    \Input{vector of average training loss values ($\mathcal{L}$), real training data features ($X_{train}$), real training data labels ($Y_{train}$), synthetic training data features ($X_{syn\_train}$), pseudo labels for training ($Y_{syn\_train}$), upper-level percentile ($p_{upper}$), lower-level percentile ($p_{lower}$), class labels ($L$)}
    \Output{selected data features ($X_{select}$), selected data labels ($Y_{select}$)} 
    \BlankLine
    $X$ = \texttt{concatenate}($X_{train}$, $X_{syn\_train}$)\;
    $Y$ = \texttt{concatenate}($Y_{train}$, $Y_{syn\_train}$)\;
    \ForEach {$l \in L$}{
        $\text{set upper-level threshold } t_{upper}^l = \text{the } p_{upper} \text{-th percentile in } \mathcal{L}~|~l$\;
        $\text{set lower-level threshold } t_{lower}^l = \text{the } p_{lower} \text{-th percentile in } \mathcal{L}~|~l$\;
        $\text{select } (X_{select}, Y_{select})~|~l \text{ from } \lbag t_{lower}^l \leq \text{data in } (X, Y)~|~l$
        $\text{ with average training loss values in } \mathcal{L}~|~l \leq t_{upper}^l \rbag$\;
        $X_{select} = \texttt{concatenate} (X_{select}, X_{select}~|~l)$\;
        $Y_{select} = \texttt{concatenate} (Y_{select}, Y_{select}~|~l)$\;
    }
    \caption{Data Selection} 
    \label{alg:DS}
\end{algorithm}

Algorithm \ref{alg:DS} shows the pseudocode of the DS module. It starts by concatenating all training data, including 
the real training data ($X_{train}$, $Y_{train}$) and the synthetic training data ($X_{syn\_train}$, $Y_{syn\_train}$). 
Then, for each class label $l$, the module sets an upper-level threshold $t_{upper}^l$ and a lower-level threshold 
$t_{lower}^l$ as the $p_{upper}$-th and the $p_{lower}$-th percentile of the average training loss values in 
$\mathcal{L}~|~l$, respectively. Next, the algorithm selects the data ($X_{select}$, $Y_{select}$)$~|~l$ from the 
bag where data in ($X, Y$)$~|~l$ have average training loss values in $\mathcal{L}~|~l$ that fall between the
values of their corresponding $t_{lower}^l$ and $t_{upper}^l$. Finally, the selected data ($X_{select}$,
$Y_{select}$) are retrieved by concatenating all ($X_{select}$, $Y_{select}$)$~|~l$. By doing so, the DS module selects 
the data for the extended calibration set in a stratified manner.

\subsubsection{Non-Conformity Measure}
\label{non-con}
Next, we calculate the non-conformity scores of the data in our extended calibration set, as described in 
Section \ref{sec:ICP}. The extended calibration set includes the real validation data from the new domain, the 
synthetic data for validation, and the selected data from Process 2. We use their output responses from the 
updated model as inputs to the non-conformity measure. Next, we need to define a measure that captures 
the non-conformity of a prediction precisely. 

Given a set of all possible class labels $y^1, y^2, ..., y^c$, the softmax output response of a model for a data
feature is a probability distribution vector [$o^1, o^2, ..., o^c$] that corresponds to the probability
distribution of its true label among $y^1, y^2, ..., y^c$, where $\sum_{i=1}^c o^i = 1$. Therefore, for a data
feature with a true class label $y^j~|~j = 1, ..., c$, the higher the output probability $o^j$ of the true label, the more conforming the prediction is. On the contrary, the higher the other output probabilities $o^i~|~i \neq j$, the less conforming the prediction is. Consequently, the most important output probability is the one with the maximum value $\max(o^i)~|~i = 1, ..., c;~i \neq j$ except for the true label $o^j$ since it is the one that signifies model uncertainty. 

We modify the function from \cite{ICP4NN} to define our non-conformity measure for a data feature with a provisional class label assignment $y^j$ as:

\begin{equation}
    \label{eq:3}
    \alpha^{(j)} = \frac{\max(o^i)~|~i = 1, ..., c;~i \neq j}{o^j \times \gamma},
\end{equation}

\noindent where $\gamma > 0$. Parameter $\gamma$ enables us to gain control over the sensitivity of the resulting 
non-conformity score based on the importance of the underlying output probability. By increasing $\gamma$, we reduce the 
importance of $o^j$ and increase the importance of all other output probabilities. Accordingly, we use
Eq.~(\ref{eq:3}) to compute the non-conformity scores of all data in our extended calibration set in this process. 

\subsubsection{p-Value Function}
\label{sec:p-func}
In this last process of EICP, we perform prediction for the given new data instance. Similar to Process 2, we pass 
the new data instance through the updated model to retrieve its output response. Then, we use its output probability 
distribution to compute its non-conformity score. Finally, we use the non-conformity scores from the extended 
calibration set and the new data instance to perform prediction with a p-value function. Based on the calculated 
p-values, we can produce a prediction result, confidence score, and credibility value for the new data instance. 

Let our extended calibration set be $Z_{calibrate}$ with $q$ data instances. For a given new data instance 
$z_{new} = (x_{new}, y_{new})$ with possible class labels $y^1, y^2, ..., y^c$, we compute the p-value of a 
provisional class label assignment $y^j~|~j = 1, ...,c$ as:

\begin{equation*}
    p(z_{new}^{(j)}) := \frac{\# \{\alpha_{new}^{(j)} \leq \alpha_{i}~|~\forall i \in \lbag Z_{calibrate}, z_{new}^{(j)} \rbag \}}{q+1}.
\end{equation*}

\noindent After computing all p-values for all possible $y^j$'s, we predict the class label of the new data instance 
as the one that yields \emph{the largest} p-value. It can be written as:

\begin{equation*}
    y_{new} := \argmax\limits_{y^j} p(z_{new}^{(j)}).
\end{equation*}

\noindent Finally, for this prediction result, we output the confidence score as one minus \emph{the second largest} 
p-value and the credibility value as \emph{the largest} p-value. They are given as:

\begin{equation*}
    \text{confidence score} := 1 - \text{the second largest } p(z_{new}^{(j)}),
\end{equation*}
\begin{equation*}
    \text{credibility value} := \max(p(z_{new}^{(j)})~|~j = 1, ..., c).
\end{equation*}

\section{Experimental Setup}
\label{sec:setup}

We describe the experimental setup next.

\subsection{Datasets}
\label{sec:datasets}
We have chosen three different disease datasets collected from commercial WMSs to evaluate PAGE's effectiveness in 
domain-incremental adaptation for disease detection. These are the CovidDeep \cite{covidD}, DiabDeep \cite{diabD}, 
and MHDeep \cite{mhD} datasets from the literature. Data collection and experimental procedures for these datasets 
were governed by Institutional Review Board approval. The efficacy of these datasets has been validated by results 
presented in the literature \cite{covidD, diabD, mhD, doctor}.

Table~\ref{tbl:CDFeatures} shows the data features collected in the CovidDeep dataset. It contains various continuous 
physiological signals and Boolean responses to a simple questionnaire. The dataset was collected from 38 healthy 
individuals, 30 asymptomatic patients, and 32 symptomatic patients at San Matteo Hospital in Pavia, Italy. The data 
were acquired using commercial WMSs and devices, including an Empatica E4 smartwatch, a pulse oximeter, and a blood 
pressure monitor. Hassantabar \textit{et al.} conducted numerous experiments \cite{covidD} to identify which data 
features are the most beneficial to detecting COVID-19 in patients for obtaining a high disease-detection accuracy. 

Table~\ref{tbl:DDFeatures} shows the data features recorded in the DiabDeep dataset. It includes the physiological 
signals of participants and additional electromechanical and ambient environmental data. All data features have
continuous values. The dataset was obtained from 25 non-diabetic individuals, 14 Type-I diabetic patients, and 13 
Type-2 diabetic patients. While the physiological data were collected with an Empatica E4 smartwatch, other data 
were recorded using a Samsung Galaxy S4 smartphone. The additional smartphone data were proven to provide diagnostic 
insights through user habit tracking \cite{additional_data}. For instance, they help with body movement tracing and 
physiological signal calibration. 

Table~\ref{tbl:DDFeatures} also shows the data features recorded in the MHDeep dataset. It collects the same physiological 
signals and additional data features as those collected in the DiabDeep dataset. The dataset was obtained from 
23 healthy participants, 23 participants with bipolar disorder, 10 participants with major depressive disorder, and 
16 participants with schizoaffective disorder at the Hackensack Meridian Health Carrier Clinic, Belle Mead, New Jersey. 
As before, the data were collected using an Empatica E4 smartwatch and a Samsung Galaxy S4 smartphone. 

\begin{table}[t]
    \caption{Data Features Collected in the CovidDeep Dataset}
    \label{tbl:CDFeatures}
    \centering
    \resizebox{0.65\linewidth}{!}{
    \begin{tabular}{lcc}
    \toprule
    Data Features & Data Source & Data Type \\
    \midrule
    Galvanic Skin Response ($\mu S$) & \\
    Skin Temperature ($^\circ C$) & Smartwatch & Continuous \\
    Inter-beat Interval ($ms$) & \\
    \midrule
    Oxygen Saturation ($\%$) & Pulse Oximeter & Continuous \\
    Systolic Blood Pressure (mmHg) & Blood Pressure Monitor & Continuous \\
    Diastolic Blood Pressure (mmHg) & Blood Pressure Monitor & Continuous \\
    \midrule
    Immune-compromised & \\
    Chronic Lung Disease & \\
    Shortness of Breath & \\
    Cough & \\
    Fever & \\
    Muscle Pain & Questionnaire & Boolean \\
    Chills & \\
    Headache & \\
    Sore Throat & \\
    Smell/Taste Loss & \\
    Diarrhea & \\
    \bottomrule
    \end{tabular}
    }
\end{table}

\begin{table}[t]
    \caption{Data Features Collected in the DiabDeep and MHDeep Datasets}
    \label{tbl:DDFeatures}
    \centering
    \resizebox{0.55\linewidth}{!}{
    \begin{tabular}{lcc}
    \toprule
    Data Features & Data Source & Data Type \\
    \midrule
    Galvanic Skin Response ($\mu S$) & \\
    Skin Temperature ($^\circ C$) & \\
    Acceleration ($x, y, z$) & Smart Watch & Continuous \\
    Inter-beat Interval ($ms$) & \\
    Blood Volume Pulse & \\
    \midrule
    Humidity & \\
    Ambient Illuminance & \\
    Ambient Light Color Spectrum & \\
    Ambient Temperature & \\
    Gravity ($x, y, z$) & \\
    Angular Velocity ($x, y, z$) & \\
    Orientation ($x, y, z$) & Smart Phone & Continuous \\
    Acceleration ($x, y, z$) & \\
    Linear Acceleration ($x, y, z$) & \\
    Air Pressure & \\
    Proximity & \\
    Wi-Fi Radiation Strength & \\
    Magnetic Field Strength & \\
    \bottomrule
    \end{tabular}
    }
\end{table}

\subsection{Dataset Preprocessing}
\label{sec:preprocessing}
First, we preprocess all datasets to transform them into a suitable format for our experiments. We synchronize and 
window the data streams by dividing data into 15-second windows with 15-second shifts in between to avoid time 
correlation between adjacent data windows. Each 15-second window of data constitutes one data instance. Next, we 
flatten and concatenate the data within the same time window from the WMSs and smartphones. Then, we concatenate 
the sequential time series data with responses to the questionnaire for the CovidDeep dataset. This results in a 
total of 14047 data instances with 155 features each for the CovidDeep dataset, a total of 20957 data instances with 
4485 features each for the DiabDeep dataset, and a total of 27082 data instances with 4485 features each for the 
MHDeep dataset. Subsequently, we perform min-max normalization on the feature data in all datasets to scale them into 
the range between 0 and 1. This prevents features with a wider value range from overshadowing those with a narrower range.

Next, we perform dimensionality reduction on the DiabDeep and MHDeep datasets using principal component analysis to 
reduce their dimension from 4485 to 155. First, we standardize the feature data in the datasets and compute the 
covariance matrix of the features. Then, we perform eigendecomposition to find the eigenvectors and eigenvalues of the 
covariance matrix to identify the principal components. Next, we sort the eigenvectors in descending order based on the 
magnitude of their corresponding eigenvalues. Finally, we construct a projection matrix to select the top 155 principal 
components. Afterward, we recast the data along the principal component axes to retrieve the 155-dimensional feature 
data for these datasets. 

Due to the limited amount of patient data available in the datasets, we assume two different data domains in our 
domain-incremental adaptation experiments. For each dataset, we first randomly split the patients into two domains in 
a stratified fashion, where Domain 1 has 80\% of the patients from each class and Domain 2 has the remaining 20\%. Next, 
within each domain, we take the first 70\%, the next 10\%, and the last 20\% of each patient's sequential time series 
data to construct the training, validation, and test sets with no time overlap. Finally, we apply the Synthetic Minority 
Oversampling Technique (SMOTE) \cite{SMOTE} to the partitioned training datasets to counteract the data imbalance 
issue within each dataset. We start by selecting a random data instance $A$ in a minority class and finding its five 
nearest neighbors in that class. Then, we randomly select one nearest neighbor $B$ from the five and draw a line segment 
between $A$ and $B$ in the feature space. Finally, we generate a synthetic data instance at a randomly selected point 
on the line segment between $A$ and $B$. We then repeat this process until we obtain a balanced number of data instances 
in all classes in each partitioned training set.

\subsection{Implementation Details}
\begin{figure}[t]
    \centering
    \includegraphics[width=0.6\linewidth]{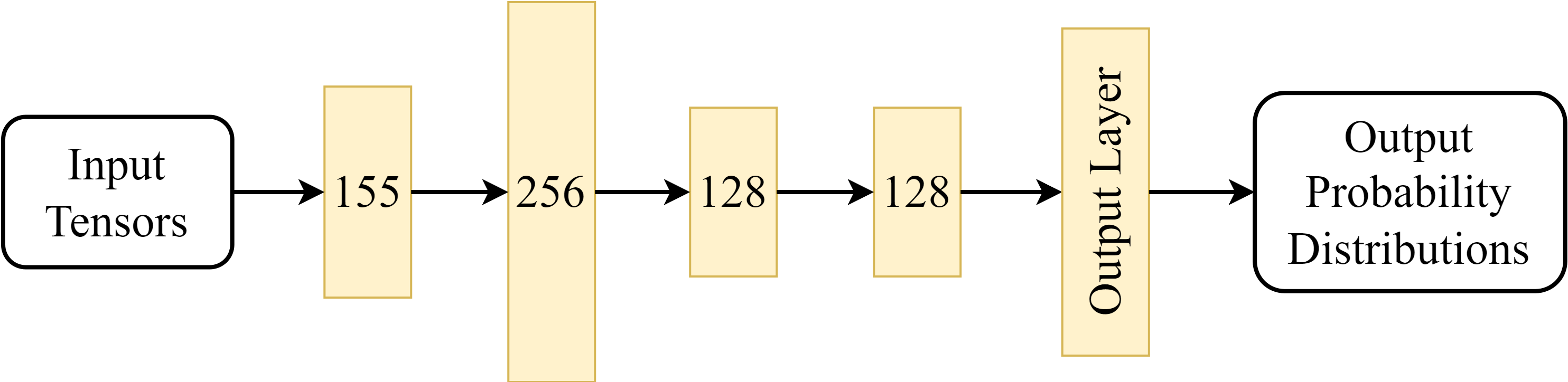}
    \caption{The DNN architecture used in experiments.}
    \label{fig:dnn}
\end{figure}

PAGE is applicable to any generic DNN model that performs classification on sequential time series data with a softmax 
output layer. For experimental simplicity, we use an MLP model to evaluate our PAGE strategy. We use grid search to 
obtain the best hyperparameter values for our MLP model, using the validation sets of the three datasets. We find 
that a four-layer architecture provides the optimal performance, in general. Fig.~\ref{fig:dnn} shows the architecture 
of the final MLP model. It has an input layer with 155 neurons to align with the input dimensionality of the datasets. 
Subsequently, it has three hidden layers with 256, 128, and 128 neurons in each layer, respectively. Finally, it has 
an output layer with the number of neurons that corresponds to the number of possible classes. We use the rectified 
linear unit (ReLU) as the nonlinear activation function in the hidden layers and the softmax function for the output 
layer to generate the output probability distributions.

We use the stochastic gradient descent optimizer with a momentum of 0.9 in our experiments and initialize the learning 
rate to 0.005. We set the batch size to 128 for training, where data are drawn evenly from the real and synthetic 
training data, as described in Section \ref{sec:MU}. We train the MLP model for 300 epochs. We set the maximum number 
of Gaussian model components ($C_{max}$) for the SDG module to 10. In addition, based on the ablation study 
described in Section \ref{sec:ablation2}, we set the number of synthetic data instances ($count$) to 90\% of the 
number of real data features for training GMM ($X_{train\_gmm}$) in the new domain. In other words, the number of 
synthetic training (validation) data instances is 90\% of the number of the new real training (validation) data 
instances.  Similarly, based on the ablation study described in Section \ref{sec:ablation3}, we set the upper-level 
percentile ($p_{upper}$) to the 90th percentile and the lower-level percentile ($p_{lower}$) to the 70th percentile 
of the average training loss values of each classification class in the DS module in EICP. Finally, we set the
parameter $\gamma$ in Eq.~(\ref{eq:3}) to 2.0 for our non-conformity measure. 

We implement PAGE with PyTorch and perform the experiments on an NVIDIA A100 GPU. We employ CUDA and cuDNN libraries 
to accelerate the experiments. 

\section{Experimental Results}
\label{sec:results}

We present experimental results next.

\begin{table*}[t]
    \caption{Domain-incremental Adaptation Experimental Results}
    \label{tbl:DIA}
    \centering
    \resizebox{1\linewidth}{!}{
    \begin{tabular}{cccccccc}
        \toprule
        Datasets & Frameworks & Domain 1 $Acc$ & Domain 2 $Acc$ & $Acc_{avg}$ & F1-score$_{avg}$ & BWT & Buffer Size (MB) \\
        \midrule
         & w/o Domain Adaptation & 0.977 & 0.808 & 0.893 & 0.930 & - & - \\
         & Naive Fine-tuning & 0.767 & 1.000 & 0.884 & 0.929 & -0.217 & 0 \\
         & Ideal Joint-training & 0.992 & 0.998 & 0.995 & 0.998 & 0.008 & 5.8 \\
        CovidDeep & DOCTOR & 0.979 & 1.000 & 0.990 & 0.994 & -0.007 & 1.7 \\
         & EWC & 0.844 & 1.000 & 0.922 & 0.958 & -0.125 & 0.7 \\
         & LwF & 0.883 & 1.000 & 0.941 & 0.968 & -0.097 & 0 \\
        \cmidrule{2-8}
         & PAGE & 0.952 & 0.999 & 0.976 & 0.984 & -0.025 & 0 \\
        \midrule
         & w/o Domain Adaptation & 0.921 & 0.368 & 0.645 & 0.647 & - & - \\
         & Naive Fine-tuning & 0.445 & 1.000 & 0.723 & 0.796 & -0.477 & 0 \\
         & Ideal Joint-training & 0.920 & 0.991 & 0.955 & 0.960 & -0.002 & 11.5 \\
        DiabDeep & DOCTOR & 0.918 & 0.995 & 0.957 & 0.961 & -0.004 & 3.4 \\
         & EWC & 0.648 & 0.958 & 0.803 & 0.822 & -0.280 & 0.7 \\
         & LwF & 0.751 & 0.980 & 0.866 & 0.876 & -0.185 & 0 \\
        \cmidrule{2-8}
         & PAGE & 0.918 & 0.977 & 0.947 & 0.949 & -0.003 & 0 \\
        \midrule
         & w/o Domain Adaptation & 0.824 & 0.108 & 0.466 & 0.753 & - & - \\
         & Naive Fine-tuning & 0.280 & 0.986 & 0.633 & 0.794 & -0.554 & 0 \\
         & Ideal Joint-training & 0.834 & 0.942 & 0.888 & 0.979 & -0.006 & 13.6 \\
        MHDeep & DOCTOR & 0.819 & 0.954 & 0.887 & 0.977 & -0.013 & 4.1 \\
         & EWC & 0.434 & 0.923 & 0.678 & 0.830 & -0.428 & 0.7 \\
         & LwF & 0.577 & 0.936 & 0.757 & 0.883 & -0.262 & 0 \\
        \cmidrule{2-8}
         & PAGE & 0.782 & 0.928 & 0.855 & 0.954 & -0.050 & 0 \\
         \bottomrule
    \end{tabular}
    }
\end{table*}

\subsection{Domain-Incremental Adaptation Experiments}
\label{sec:DIA_Exp}
First, we evaluate the performance of PAGE for domain-incremental adaptation on the datasets introduced in Section 
\ref{sec:datasets}. We compare PAGE against five different strategies, including naive fine-tuning, ideal joint-training, 
DOCTOR \cite{doctor}, EWC \cite{ewc}, and LwF \cite{lwf}. The naive fine-tuning method naively fine-tunes the MLP model 
with real training data from new domains without adopting any CL algorithm. The ideal joint-training method represents 
the ideal scenario where we have access to all training data from both past and new domains to train the MLP model with 
all data jointly. DOCTOR performs replay-style CL by either preserving raw training data from past domains or using
data from previous domains to generate synthetic data for future replays. We report its best performance based on its two 
algorithms. EWC stores a Fisher information matrix for learned domains to weigh the importance of each model parameter. 
Then, it imposes a quadratic penalty on the update process of the important parameters. LwF distills the knowledge of 
learned domains with the pseudo labels given to the real training data from new domains. Then, it trains the model on 
the new real training data with both their pseudo and ground-truth labels.

Table \ref{tbl:DIA} presents the domain-incremental adaptation experimental results for all three datasets. The results 
are reported on the \emph{real} test data from both domains of all datasets. As shown in the first row for each dataset, 
the MLP model without domain adaptation performs poorly on data from the second domain and has a low average test 
accuracy. The naive fine-tuning method makes the MLP model overfit to the second domain and performs poorly on the 
first domain due to CF. PAGE outperforms the naive fine-tuning method and achieves superior results compared to EWC and 
LwF. While DOCTOR and the ideal joint-training method perform very well on domain-incremental adaptation and achieve 
high average test accuracy, they require a memory buffer that scales up with the number of new domains to preserve data 
from all seen domains. On the other hand, PAGE achieves very competitive results with negligible loss in average 
test accuracy without incurring the memory buffer cost and suffering from scalability issues. 

\begin{table*}[t]
    \caption{Conformal Prediction Results on the CovidDeep Dataset}
    \label{tbl:CovidCP}
    \centering
    \resizebox{1\linewidth}{!}{
    \begin{tabular}{c|c|c|c|c|c||c|c|c|c|c}
        \toprule
         & \textbf{ICP} & Correct & Incorrect & Correctness & Error Rate & \textbf{EICP} & Correct & Incorrect & Correctness & Error Rate \\
        \midrule
        \multirow{2}{*}{Domain1} & Certain & 696 & 0 & \multirow{2}{*}{0.330} & \multirow{2}{*}{0} & Certain & 1386 & 0 & \multirow{2}{*}{0.656} & \multirow{2}{*}{0} \\
         & Uncertain & 1227 & 189 & & & Uncertain & 537 & 189 & & \\
        \midrule
        \multirow{2}{*}{Domain2} & Certain & 243 & 0 & \multirow{2}{*}{0.349} & \multirow{2}{*}{0} & Certain & 520 & 0 & \multirow{2}{*}{0.746} & \multirow{2}{*}{0} \\
         & Uncertain & 454 & 0 & & & Uncertain & 177 & 0 & & \\
        \bottomrule
    \end{tabular}
    }
\end{table*}

\begin{table*}[t]
    \caption{Conformal Prediction Results on the DiabDeep Dataset}
    \label{tbl:DiabCP}
    \centering
    \resizebox{1\linewidth}{!}{
    \begin{tabular}{c|c|c|c|c|c||c|c|c|c|c}
        \toprule
         & \textbf{ICP} & Correct & Incorrect & Correctness & Error Rate & \textbf{EICP} & Correct & Incorrect & Correctness & Error Rate \\
        \midrule
        \multirow{2}{*}{Domain1} & Certain & 708 & 0 & \multirow{2}{*}{0.212} & \multirow{2}{*}{0} & Certain & 1321 & 0 & \multirow{2}{*}{0.395} & \multirow{2}{*}{0} \\
         & Uncertain & 2301 & 338 & & & Uncertain & 1688 & 338 & & \\
        \midrule
        \multirow{2}{*}{Domain2} & Certain & 201 & 0 & \multirow{2}{*}{0.239} & \multirow{2}{*}{0} & Certain & 445 & 0 & \multirow{2}{*}{0.529} & \multirow{2}{*}{0} \\
         & Uncertain & 662 & 18 & & & Uncertain & 378 & 18 & & \\
        \bottomrule
    \end{tabular}
    }
\end{table*}

\begin{table*}[t]
    \caption{Conformal Prediction Results on the MHDeep Dataset}
    \label{tbl:MHCP}
    \centering
    \resizebox{1\linewidth}{!}{
    \begin{tabular}{c|c|c|c|c|c||c|c|c|c|c}
        \toprule
         & \textbf{ICP} & Correct & Incorrect & Correctness & Error Rate & \textbf{EICP} & Correct & Incorrect & Correctness & Error Rate \\
        \midrule
        \multirow{2}{*}{Domain1} & Certain & 785 & 18 & \multirow{2}{*}{0.181} & \multirow{2}{*}{0.004} & Certain & 1430 & 32 & \multirow{2}{*}{0.329} & \multirow{2}{*}{0.007} \\
         & Uncertain & 2556 & 986 & & & Uncertain & 1911 & 972 & & \\
        \midrule
        \multirow{2}{*}{Domain2} & Certain & 30 & 2 & \multirow{2}{*}{0.028} & \multirow{2}{*}{0.002} & Certain & 230 & 2 & \multirow{2}{*}{0.214} & \multirow{2}{*}{0.002} \\
         & Uncertain & 961 & 82 & & & Uncertain & 761 & 82 & & \\
        \bottomrule
    \end{tabular}
    }
\end{table*}

\subsection{Extended Inductive Conformal Prediction Experiments}
\label{sec:EICP_Exp}
At test time, users can input their pre-processed WMS data into PAGE and obtain the disease-detection results, 
confidence scores, and credibility values, as shown in Fig.~\ref{fig:schematic}. The confidence scores and credibility 
values can help users determine if their detection results are robust enough and whether clinical intervention is 
required. For healthy results with high confidence scores and credibility values, clinical intervention is not
necessary. For disease-positive results with high confidence scores and credibility values, users can be alerted to 
seek medical treatments. 

To demonstrate and evaluate the performance of EICP in PAGE against the original ICP, we report their performances in 
a confusion matrix format. Tables \ref{tbl:CovidCP}, \ref{tbl:DiabCP}, and \ref{tbl:MHCP} show the CP experimental 
results for the CovidDeep, DiabDeep, and MHDeep datasets, respectively. The results are reported on the \emph{real} 
test data from both domains of all three datasets. The rows of the confusion matrices indicate whether the model is 
certain or uncertain about its detection results. We say that the model is certain when its confidence score and 
credibility value are higher than certain thresholds. We select the threshold values for each dataset based on their 
validation datasets. The confidence score threshold for all three datasets is 90\%. The credibility value thresholds 
for the CovidDeep, DiabDeep, and MHDeep datasets are 70\%, 85\%, and 90\%, respectively. The columns indicate whether 
the detection results are correct or incorrect. The detection result is correct when it correctly predicts the 
ground-truth label. We define two metrics to evaluate CP performance: correctness and error rate. Correctness is 
defined as the number of certain and correct detection results over the number of all test data instances. Error 
rate is defined as the number of certain but incorrect detection results over the number of all test data instances. 

As shown in the tables, EICP achieves superior correctness compared to the original ICP with an identical error rate 
for both CovidDeep and DiabDeep datasets. For MHDeep, EICP outperforms ICP in correctness with only a 0.3\% increase 
in the error rate for Domain 1. These results demonstrate that EICP reflects model uncertainty more precisely than ICP 
because of its smaller number of correct but uncertain results. EICP also provides a more accurate statistical 
guarantee for disease detection (correct and certain). In addition, the results show that EICP reduces clinical
workload by up to 75\% (based on the correctness metric).

\subsection{Ablation Study}
\label{sec:ablation}

Next, we present the results of some ablation studies.

\subsubsection{Probability Density Estimation Method}
\label{sec:ablation1}

\begin{table*}[t]
    \caption{Ablation Study for the Probability Density Estimation Method}
    \label{tbl:abl1}
    \centering
    \resizebox{1\linewidth}{!}{
    \begin{tabular}{cccccccc}
        \toprule
        Datasets & Frameworks & Domain 1 $Acc$ & Domain 2 $Acc$ & $Acc_{avg}$ & F1-score$_{avg}$ & BWT & Buffer Size (MB) \\
        \midrule
        \multirow{2}{*}{CovidDeep} & PAGE with GMM & 0.952 & 0.999 & 0.976 & 0.984 & -0.025 & 0 \\
         & PAGE with KDE & 0.904 & 0.981 & 0.942 & 0.970 & -0.079 & 0 \\
        \midrule
        \multirow{2}{*}{DiabDeep} & PAGE with GMM & 0.918 & 0.977 & 0.947 & 0.949 & -0.003 & 0 \\
         & PAGE with KDE & 0.689 & 0.871 & 0.780 & 0.806 & -0.239 & 0 \\
        \midrule
        \multirow{2}{*}{MHDeep} & PAGE with GMM & 0.782 & 0.928 & 0.855 & 0.954 & -0.050 & 0 \\
         & PAGE with KDE & 0.669 & 0.813 & 0.741 & 0.897 & -0.163 & 0 \\
         \bottomrule
    \end{tabular}
    }
\end{table*}

In this ablation study, we construct the SDG module with the non-parametric kernel density estimation (KDE) method. 
Then, we compare its performance in domain-incremental adaptation with the SDG module composed of the parametric 
GMM density estimation method.

For a mathematical background and introduction to KDE, see \cite{tutor,doctor}. Here, we follow DOCTOR 
\cite{doctor} to construct the KDE SDG module. We use a similar algorithm to Algorithm \ref{alg:SDG} to generate 
synthetic training data ($X_{syn\_train}, Y_{syn\_train}$) and synthetic validation data 
($X_{syn\_valid}, Y_{syn\_valid}$). 

As shown in Table~\ref{tbl:abl1}, PAGE with GMM greatly outperforms PAGE with KDE in all metrics except that they both 
do not need a memory buffer. This is the reason we chose the GMM density estimation method to construct our SDG module 
in PAGE.

\subsubsection{Amount of Synthetic Data for Replay}
\label{sec:ablation2}

\begin{figure*}[t]
    \centering
    \begin{subfigure}[b]{0.4\textwidth}
        \centering
        \includegraphics[width=\textwidth]{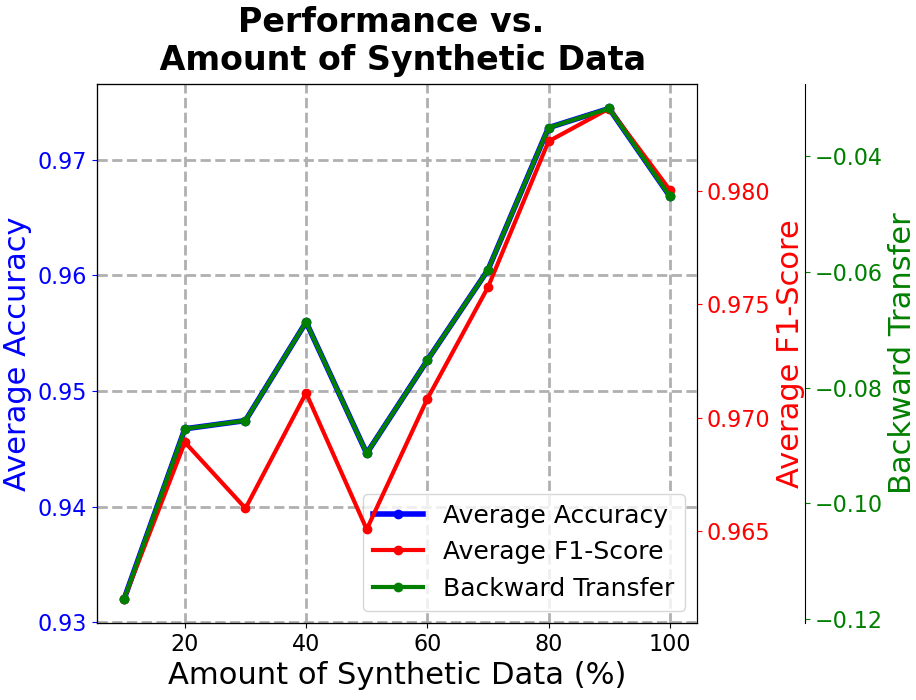}
        \caption{CovidDeep Dataset}
        \label{fig:abl2covid}
    \end{subfigure}
    \begin{subfigure}[b]{0.4\textwidth}
        \centering
        \includegraphics[width=\textwidth]{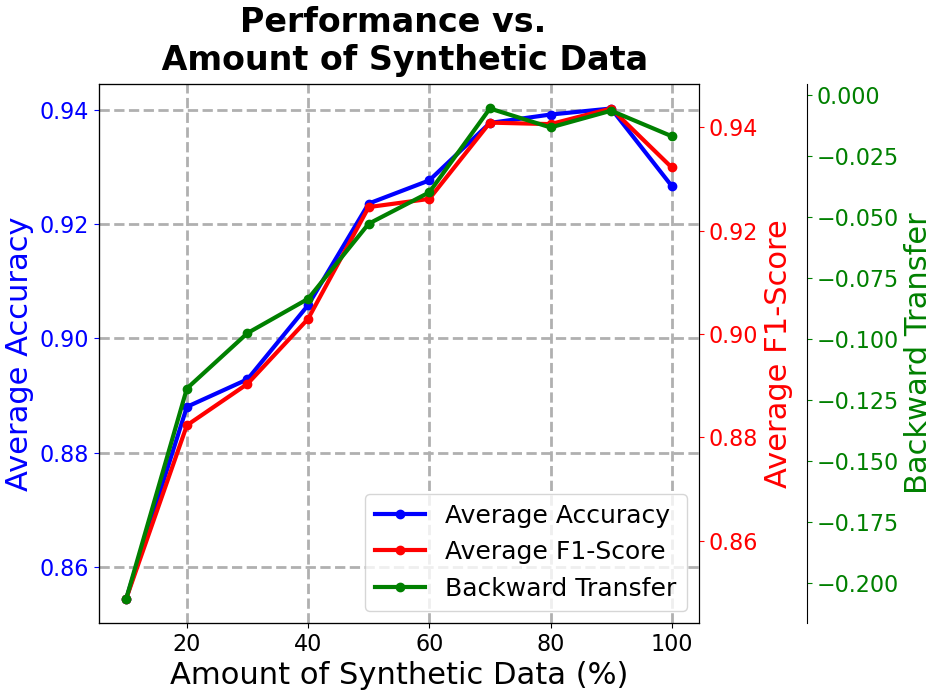}
        \caption{DiabDeep Dataset}
        \label{fig:abl2diab}
    \end{subfigure}
    \begin{subfigure}[b]{0.4\textwidth}
        \centering
        \includegraphics[width=\textwidth]{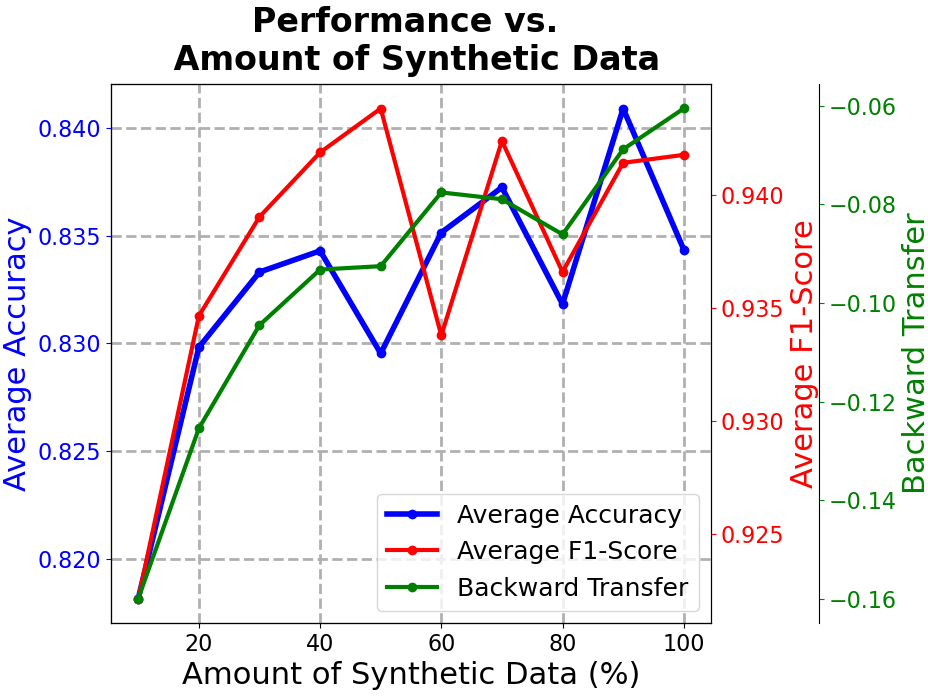}
        \caption{MHDeep Dataset}
        \label{fig:abl2mh}
    \end{subfigure}
    \caption{Domain-incremental adaptation performance vs.~the amount of synthetic data generated for the (a) CovidDeep, (b) DiabDeep, and (c) MHDeep datasets. (Best viewed in color.)}
    \label{fig:abl2}
\end{figure*}

Next, we perform an ablation study of the amount of synthetic data required for PAGE to perform domain-incremental 
adaptation. Figs.~\ref{fig:abl2covid}, \ref{fig:abl2diab}, and \ref{fig:abl2mh} show the ablation study results on 
the domain-incremental adaptation performance versus the amount of synthetic data used for replay for the three datasets. 
The $y$ axes in the figures denote curves for average accuracy, average F1-score, and BWT. The $x$ axes represent the 
amount of synthetic data as a percentage of the amount of the real data from the new domain. When the amount of synthetic 
data used for replay is around 90\% of the real data from the new domain, PAGE achieves the highest average accuracy. 
When more synthetic data are generated for replay, the chance that the model overfits to past domains and 
underperforms on the new domain becomes higher, which might impact average accuracy. Therefore, we set the number of 
synthetic data instances ($count$) in SDG to 90\% of the number of the real data instances from the new domain for 
replay in PAGE. 

\subsubsection{Data Selection Range}
\label{sec:ablation3}
\begin{figure*}[t]
    \centering
    \begin{subfigure}[b]{0.45\textwidth}
        \centering
        \includegraphics[width=\textwidth]{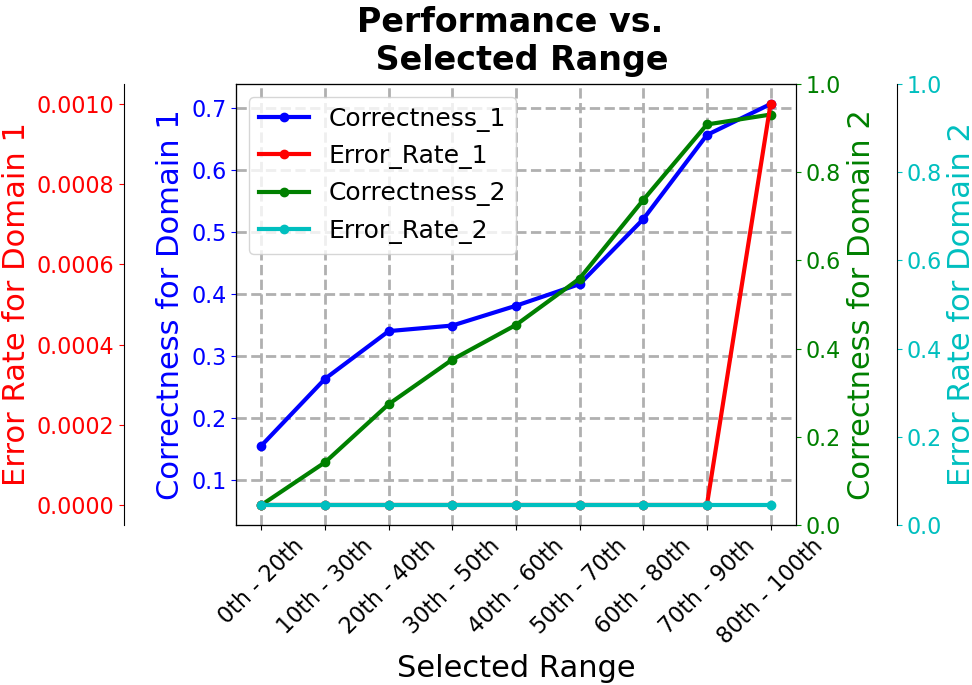}
        \caption{CovidDeep Dataset}
        \label{fig:abl3covid}
    \end{subfigure}
    \begin{subfigure}[b]{0.45\textwidth}
        \centering
        \includegraphics[width=\textwidth]{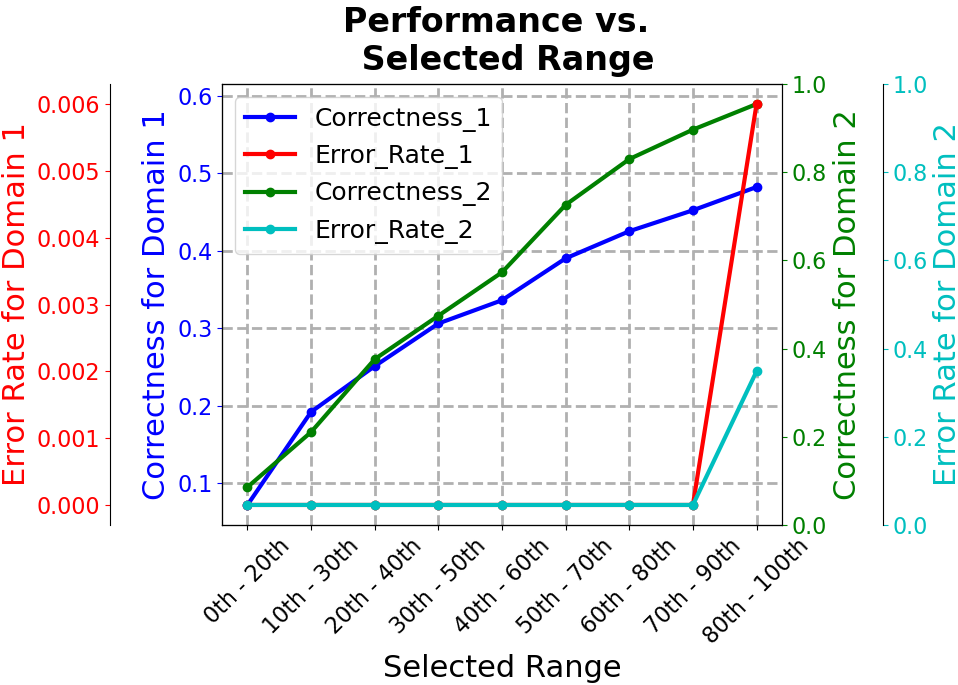}
        \caption{DiabDeep Dataset}
        \label{fig:abl3diab}
    \end{subfigure}
    \begin{subfigure}[b]{0.45\textwidth}
        \centering
        \includegraphics[width=\textwidth]{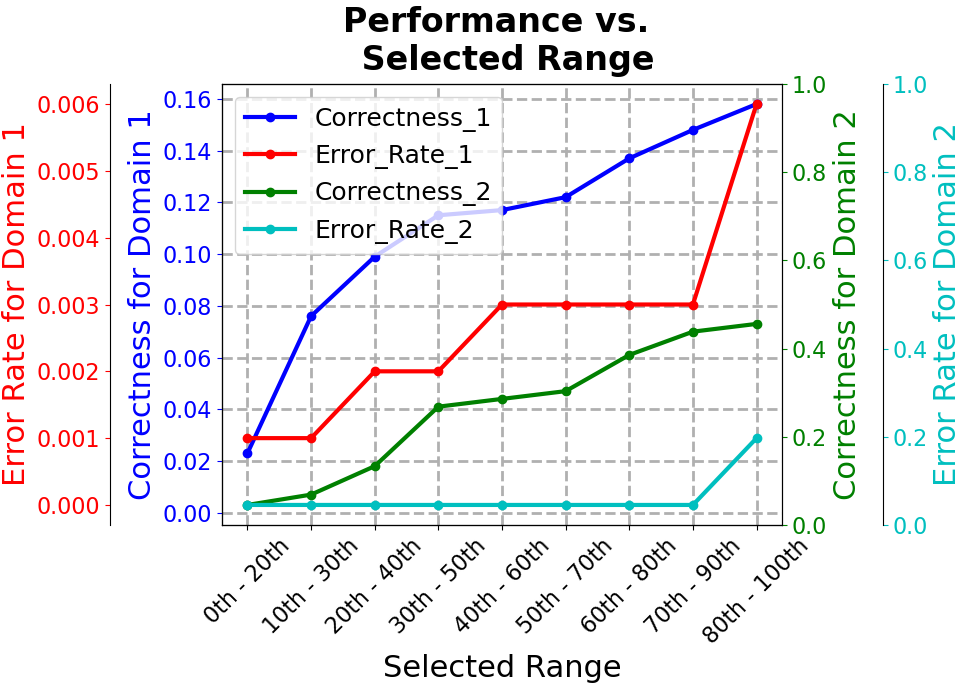}
        \caption{MHDeep Dataset}
        \label{fig:abl3mh}
    \end{subfigure}
    \caption{EICP performance vs.~the selected range for the DS module for the (a) CovidDeep, (b) DiabDeep, and (c) MHDeep datasets. (Best viewed in color.)}
    \label{fig:abl3}
\end{figure*}

In this experiment, we investigate the ranges for setting the lower-level percentile ($p_{lower}$) and the upper-level 
percentile ($p_{upper}$) for the DS module when performing EICP. As described in Section \ref{sec:DS}, the DS module 
selects the data whose average training loss values are within the range of the $p_{lower}$-th and the $p_{upper}$-th 
percentiles in the vector of the average training loss values $\mathcal{L}$. We sweep the selected range, namely the 
pair of percentiles ($p_{lower}$, $p_{upper}$), from (0th, 20th), (10th, 30th), $\ldots$, to (80th, 100th) to select the 
data from all the training data for the extended calibration set. Figs.~\ref{fig:abl3covid}, \ref{fig:abl3diab}, 
and \ref{fig:abl3mh} show the ablation study results for EICP performance versus the selected range for the DS module 
for all three datasets. We plot the correctness and error rate curves for both domains in the figures. The 
$y$ axes depict the curves for the metrics. The $x$ axes represent the selected range depicted in terms of 
($p_{lower}$, $p_{upper}$). As can be seen, EICP correctness increases for both domains as we select a higher range 
of ($p_{lower}$, $p_{upper}$) for all datasets. However, the error rates on both domains spike up in the 
$p_{lower}=80$th and $p_{upper}=100$th range for all datasets. Therefore, we set $p_{lower}$ to the 70th percentile and 
$p_{upper}$ to the 90th percentile in the DS module. 

\section{Discussions}
\label{sec:dis_lim}
The experimental results presented in Section \ref{sec:results} demonstrate that PAGE is able to perform 
domain-incremental adaptation for disease-detection tasks without the aid of any preserved data or information from 
prior domains. In addition, PAGE is able to provide model prediction interpretability and statistical guarantees
for its detection results with the help of the proposed EICP method. However, the current version of PAGE only targets 
domain-incremental adaptation scenarios. In future work, we plan to expand its scope to class-incremental and 
task-incremental scenarios for smart healthcare applications. Moreover, it will be interesting to apply PAGE to 
natural language processing and image classification tasks. 

\section{Conclusion}
\label{sec:conclusion}
We proposed PAGE, a domain-incremental adaptation strategy with past-agnostic generative replay. PAGE uses generative 
replay to perform domain-incremental adaptation while alleviating CF without the need for past knowledge in the form of 
preserved data or information. It makes PAGE highly scalable to multi-domain adaptation due to very low storage 
consumption. It also enables PAGE to protect patient privacy in disease-detection applications. Moreover, PAGE is 
directly applicable to off-the-shelf models for domain adapatation without the need for re-architecting new models 
from scratch. The proposed EICP method enables PAGE to complement prediction results with statistical guarantees and 
model prediction interpretability. This can help general users and medical practitioners determine whether further 
clinical intervention is required. 

\begin{acks}
This work was supported by NSF under Grant No. CNS-1907381.
\end{acks}

\bibliographystyle{ACM-Reference-Format}
\bibliography{biblio}

\end{document}